%% file: main.tex
\documentclass[11pt, a4paper, logo, copyright]{eai}
\usepackage{shlab}

\usepackage[authoryear, s
ort&compress, round]{natbib}
\usepackage[]{mdframed}
\newcommand{\cmark}{\textcolor{green!70!black}{\ding{51}}} 
\newcommand{\xmark}{\textcolor{red}{\ding{55}}}            
\usepackage{dblfloatfix}
\input{math_commands.tex}

\usepackage{listings}
\usepackage{hyperref}
\usepackage{url}
\usepackage{amsmath} %
\usepackage{mathrsfs} %
\usepackage{etoolbox}
\usepackage{cleveref}
\usepackage{wrapfig} 

\usepackage{tcolorbox}
\usepackage{colortbl}
\usepackage{booktabs}       %
\usepackage{amsfonts}       %
\usepackage{nicefrac}       %
\usepackage{microtype}      %
\usepackage{subcaption}
\usepackage{caption}
\usepackage{subcaption}
\usepackage{xspace} %
\usepackage{subcaption}
\usepackage{algorithm}
\usepackage{algorithmic}
\usepackage{multirow}
\usepackage{lipsum}
\usepackage{tabularx}

\usepackage{booktabs} %
\usepackage{enumitem}%
\setlist[itemize]{noitemsep, topsep=0pt}
\usepackage{enumitem,kantlipsum}

\usepackage{booktabs}
\usepackage{multirow}   
\usepackage{graphicx}    
\usepackage{pifont}
\usepackage{minted}
\usepackage{ulem}
\usepackage{longtable}
\usepackage{booktabs}
\usepackage{wrapfig}
\usepackage[table]{xcolor}
\newcommand{\modify}[1]{#1}
\usepackage{makecell}
\usepackage{listings}
\usepackage{xcolor}
\usepackage[T1]{fontenc}

\usepackage{adjustbox}
\lstdefinestyle{yamlstyle}{
    basicstyle=\ttfamily\footnotesize,
    numbers=left,
    numberstyle=\tiny,
    stepnumber=1,
    numbersep=6pt,
    frame=single,
    framerule=0.5pt,
    breaklines=true,
    breakatwhitespace=true,
    tabsize=2,
    captionpos=b,
    keywordstyle=\color{blue},
    commentstyle=\color{gray},
    stringstyle=\color{teal},
}

\newlength\savewidth

\definecolor{baselinecolor}{HTML}{d6eaf8}

\definecolor{mygray}{gray}{0.4}

\AtBeginEnvironment{tcolorbox}{\tiny}

\newcount\Comments  %
\usepackage{color}
\definecolor{darkred}{rgb}{0.9,0,0}
\definecolor{darkgreen}{rgb}{0,0.5,0}
\definecolor{darkblue}{rgb}{0,0,0.7}
\definecolor{purple}{rgb}{.6, 0,.6}
\definecolor{orange}{rgb}{1.0,0.64,0}
\newcommand{\kibitz}[2]{\ifnum\Comments=1\textcolor{#1}{#2}\fi}

\title{RoboInter1.5: A Holistic Intermediate
Representation Suite \\ for Embodied World Modeling and Robotic Manipulation}

\author[]{Team of RoboInter1.5, full author list in \hyperref[sec:contributors]{Contributors} section.}

\renewcommand{\today}{}

\begin{document}

\let\oldaddcontentsline\addcontentsline
\renewcommand{\addcontentsline}[3]{}
\input{section/0_abs}
\maketitle
\begin{figure*}[h]
    \centering
    \includegraphics[width=\linewidth]{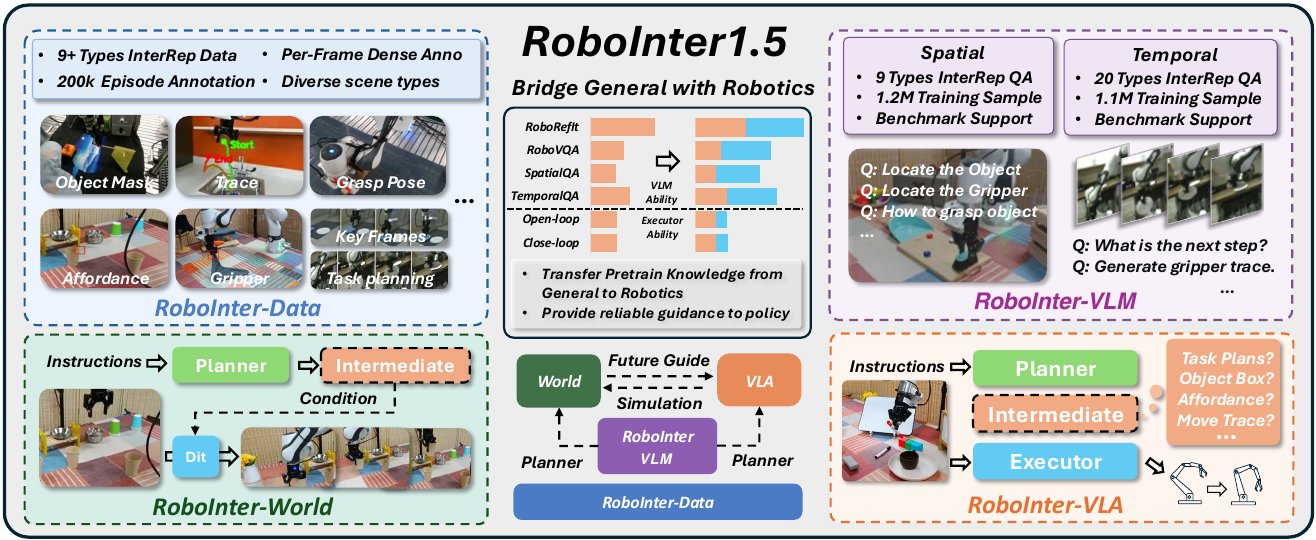}
    \caption{\textbf{RoboInter1.5 manipulation suite} includes annotation tools, annotated data, a curated VQA dataset, and their applications in VLMs, VLAs, and World Models. RoboInter provides a dataset with over 230k episodes and 10+ types of intermediate representation annotations, named \textit{RoboInter-Data}; VLMs trained on the curated embodied VQA dataset covering 29 spatial- and temporal-level categories, named \textit{RoboInter-VLM}; an integrated \textit{plan-then-execute} framework for training VLA models, \textit{RoboInter-VLA}; and an intermediate representation guided world model, \textit{RoboInter-World}.}
    \label{fig:robointer teaser}
\end{figure*}
\input{section/1_intro}
\input{section/2_related_work}

\input{section/3_dataset}
\input{section/4_method}
\input{section/5_exp}

\input{section/6_conclusion}

\bibliography{iclr2026_conference}
\clearpage
\newpage
\appendix
\input{section/supp}

\end{document}

%% file: math_commands.tex
\usepackage{amsmath,amsfonts,bm}
\usepackage{multirow}
\usepackage{subcaption}
\usepackage{wrapfig}

\def\eqref#1{equation~\ref{#1}}

\def\1{\bm{1}}

\DeclareMathAlphabet{\mathsfit}{\encodingdefault}{\sfdefault}{m}{sl}
\SetMathAlphabet{\mathsfit}{bold}{\encodingdefault}{\sfdefault}{bx}{n}

%% file: section/0_abs.tex
\begin{abstract}

Recent breakthroughs in foundation models have driven rapid progress in both vision-language-action (VLA) policies and embodied world models (WM). 
Yet, existing robot datasets remain expensive to curate, embodiment-specific, and insufficiently annotated with the fine-grained structure required for generalizable reasoning, execution, or long-horizon environment dynamics simulation.
Building on our prior work, RoboInter1.0, we present \textit{RoboInter1.5}, an extended and holistic suite of intermediate representations for both robotic manipulation and embodied world modeling. 
\textit{RoboInter1.5} provides a unified resource of data, benchmarks, and models centered on dense manipulation-oriented intermediate representations. 
Specifically, \textit{RoboInter-Data} contains over 230k manipulation episodes across 571 scenes with dense per-frame annotations covering more than ten types of intermediate representations, including subtasks, primitive skills, object and gripper grounding, segmentation, affordance, grasp poses, contact points, motion traces, etc. Built upon these annotations, 
\textit{RoboInter-VQA} introduces spatial and temporal embodied VQA tasks to benchmark and improve the intermediate-representation reasoning capabilities of our RoboInter-VLM.
\textit{RoboInter-VLA} further studies how such representations benefit action execution through implicit, explicit, and modular plan-then-execute paradigms. 
To better model the physical world, we further introduce \textit{RoboInter-World}, which leverages intermediate representations as structured conditioning signals for controllable prediction of future world states.
Extensive evaluations demonstrate that \textit{RoboInter1.5} provides a unified spatiotemporal scaffolding for intermediate representations. Rather than treating intermediate representations merely as interpretable signals, RoboInter1.5 conceptualizes them as a bidirectional interface that both regularizes low-level action spaces and constrains the latent rollouts of open-world physical simulators.

\vspace{8pt}
\links{
  \link{code}{Code:RoboInter1.5}{https://github.com/InternRobotics/RoboInter}, 
  \link{data}{Data:RoboInter-Data}{https://huggingface.co/datasets/InternRobotics/RoboInter-Data}, 
  \link{homepage}{Homepage}{https://lihaohn.github.io/RoboInter.github.io/}
}

\end{abstract}

%% file: section/1_intro.tex
\section{Introduction}
\label{sec_intro}
Recent advances in large vision-language models (VLMs) have stimulated growing interest in embodied intelligence systems, spanning embodied question answering for scene and task understanding~\citep{robobrain2.0, luo2025robobench, zhou2025roborefer, fang2025robix}, vision-language-action (VLA) models~\citep{RT-2, openvla, pi_0, bjorck2025gr00t,Vla-rl} for robotic control, and embodied world models~\citep{hung2025nora, guo2025ctrl, kim2026cosmos, ye2026world, team2026gigabrain} for simulating how the world evolve under physical interactions. While web-scale multimodal data equips VLMs with broad semantic reasoning, embodied VLA policies and generative world models struggle to directly inherit these capabilities. Meanwhile, attempting to bridge this divide by curating large-scale robot datasets~\citep{open_x_embodiment, khazatsky2024droid, wu2024robomind, bu2025agibot} remains prohibitively expensive and tightly coupled to specific embodiments~\citep{Mobile_aloha}, imposing a significant bottleneck on scalable generalization.

To address this generalization gap, recent research increasingly leverages structured intermediate representations. On the control side, \textbf{modular VLAs}~\citep{voxposer, rth, copa, moka, google2024pivot} and many \textbf{end-to-end VLAs}~\citep{zhou2025chatvla, yang2025instructvla, ecot, hirobot, onetwovla, deng2025graspvla, worldvla, wu2025momanipvla, unipi} integrate subtasks, grounding, or motion traces to decompose monolithic policies and bridge high-level reasoning with low-level control. Concurrently, broader advances in \textbf{controllable video generation}~\citep{li2024dispose, li2025magicmotion, kuang2024collaborative, chu2025wan, liang2024movideo} have demonstrated that structural priors (such as traces or keypoints) are essential to transition from unconstrained visual hallucinations to reliable dynamics synthesis. This insight has driven the adoption of these representations as structural constraints to facilitate long-horizon embodied world modeling. Ultimately, both domains converge on a shared paradigm: leveraging intermediate representations as a universal interface that seamlessly connects semantic understanding, physical execution, and future simulation.

The effectiveness of this shared paradigm critically depends on high-quality intermediate representations. Existing datasets~\citep{open_x_embodiment, khazatsky2024droid,bu2025agibot} typically pair visual inputs with high-level instructions and low-level actions, yet they rarely provide the fine-grained structural annotations for intermediate representations, which are required to bridge reasoning, control, and simulation. Furthermore, collecting and manually annotating new data from scratch is prohibitively costly and infrastructure-intensive, leaving massive community-driven open-source datasets underutilized for large-scale intermediate representations training. While recent efforts~\citep{li2025hamster, yuan2024robopoint} have explored automated annotation for existing datasets, they often encounter significant limitations. For instance, LLARVA~\citep{niu2024llarva} leverages a pretrained gripper detector to extract large-scale motion traces, but it remains highly sensitive to distribution shifts. ECoT~\citep{ecot} utilizes Gemini~\citep{GoogleDeepMind_Gemini2023} to generate pseudo-labels for textual planning and object grounding, yet struggles with precise physical alignment. ShareRobot~\citep{ji2025robobrain} combines automated pipelines with manual verification, but operates at a limited scale and yields labels that are temporally misaligned with step-wise actions. Ultimately, the absence of large-scale, high-quality annotations in current open-source datasets severely bottlenecks the advancement of intermediate representations for both VLA models and embodied world models.

To address this gap, based on our prior work, RoboInter1.0~\citep{li2026robointer}, we further introduce the \textbf{RoboInter1.5} Manipulation Suite, illustrated in Figure~\ref{fig:robointer teaser}. Built upon \textbf{RoboInter-Tool}, a lightweight GUI for the semi-automatic per-frame annotation of embodied videos, we curate \textbf{RoboInter-Data}, a large-scale dataset featuring dense intermediate representations for robotic manipulation. 
As shown in Table~\ref{tab:dataset comparison}, \textit{RoboInter-Data} encompasses over 230k episodes across 571 distinct scenes, significantly surpassing LLARVA~\citep{niu2024llarva}, ECoT~\citep{ecot}, and ShareRobot~\citep{ji2025robobrain} in both scale and environmental diversity. Unlike prior datasets constrained by limited scene variations~\citep{rh20tp} or heavily reliant on noisy automated pipelines~\citep{li2025hamster, ecot}, \textit{RoboInter-Data} uniquely guarantees high fidelity by combining scalable automatic generation with rigorous human-in-the-loop verification. This constitutes a large open-source real-world manipulation dataset offering dense, per-frame annotations across a comprehensive taxonomy of more than ten categories. These seamlessly integrate high-level semantics (subtasks, primitive skills), 2D/3D visual grounding (segmentation masks, gripper bounding boxes, affordances, placement proposals), and precise physical geometries (grasp poses, motion traces, contact points). Crucially, all annotations are strictly synchronized temporally with executed actions, robot states, and two-view visual observations (third-person and wrist-mounted cameras), establishing a robust data foundation for both end-to-end action learning and world modeling.

\input{tables/table_Compare_Other_Dataset}

Leveraging these fine-grained annotations, we develop three downstream frameworks to validate the utility of our intermediate representations systematically. First, we introduce \textbf{RoboInter-VQA} to benchmark and enhance the spatial-temporal reasoning capabilities of VLMs, comprising 9 spatial and 20 temporal VQA categories tailored for embodied scenes. Built upon the \textbf{RoboInter-VLM} planner trained on this curated VQA data, we present \textbf{RoboInter-VLA}, an integrated control framework supporting both modular and end-to-end variants. This enables rapid adaptation from semantic planning to low-level execution, allowing us to systematically investigate the impact of intermediate representations on policy generalization and controllability. Extending beyond reasoning and control, we introduce \textbf{RoboInter-World}, a controllable embodied world model. By utilizing our dense, per-frame intermediate representations as structural constraints, it effectively mitigates unconstrained visual hallucinations, enabling reliable, long-horizon forward dynamics simulation. Through extensive experiments, we demonstrate that \textit{RoboInter-Data} substantially improves the reasoning capabilities of VLM planners, particularly in understanding and generating manipulation-oriented representations. Furthermore, open- and closed-loop evaluations confirm that these structured priors provide significant performance and generalization gains to VLA policies. Meanwhile, we find that the future images predicted by \textit{RoboInter-World} can effectively improve the action prediction accuracy of \textit{RoboInter-VLA}. By analyzing the trade-offs among different VLA and world model variants, we establish a unified foundation for leveraging these data in diverse tasks. In this work, we extend several key improvements over the prior RoboInter1.0 version~\citep{li2026robointer}, including:
\begin{itemize}
  \item \textbf{Extension of the Benchmark}: Building upon \textit{RoboInter-Data}, we construct \textit{RoboInter-CV}, a novel long-horizon benchmark for embodied world modeling conditioned on intermediate representations. To the best of our knowledge, this is the first dataset in the Embodied AI field specifically designed to utilize intermediate representations as the primary guidance for world models in long-horizon manipulation tasks.
  \item \textbf{Methodological Advancement}: Extending the application of intermediate representations from VQA (\textit{RoboInter-VLM}) and Robotic Manipulation (\textit{RoboInter-VLA}) to the realm of video generation, we introduce \textit{RoboInter-World}. This novel world model architecture is explicitly designed for intermediate representation guidance. By effectively capturing and utilizing these visual cues, \textit{RoboInter-World} achieves superior generalization capabilities across diverse manipulation scenarios.
  \item \textbf{Extended Comprehensive Evaluation}: Utilizing the newly established \textit{RoboInter-CV} benchmark, we have conducted extensive new experiments to evaluate world models and how world models guide VLA models. These evaluations, including various model variants and systematic ablation studies, provide deep insights into how intermediate representations fundamentally enhance performance and physical realism.
\end{itemize}

%% file: tables/table_Compare_Other_Dataset.tex
\begin{table*}[t]
\centering
\caption{\textbf{Comparison of embodied annotation datasets.}
\textit{Emb.-VQA} denotes the availability of curated embodied VQA benchmarks and datasets; \textit{E2E-ACT} indicates whether the annotation is temporally aligned with executed actions; \textit{Curated-CoT} specifies multi-intermediate chain-of-thought support; \textit{IR-Video} indicates whether the dataset provides high-quality video sequences conditioned on dense intermediate representations}\label{tab:dataset comparison}
\begin{adjustbox}{width=\linewidth}
\scriptsize
\renewcommand{\arraystretch}{1.1}
\setlength{\tabcolsep}{3pt}
\begin{tabular}{l|ccc|cccc|cccccc|c}
\toprule
Dataset & \#Video & \#Scene & Dense & \makecell{\textbf{Emb.}\\\textbf{-VQA}} & \makecell{\textbf{E2E}\\\textbf{-ACT}} & \makecell{\textbf{Curated}\\\textbf{-CoT}} & \makecell{\textbf{IR}\\\textbf{-Video}} & \makecell{Subtask\\\&Skill} & \makecell{Affor-\\-dance} & \makecell{Contact\\Point} & \makecell{Gripper\\Box} & \makecell{Object\\Box} & Trace & \makecell{Annotation\\Type}\\
\midrule
LLARVA        & --    & 311 & \xmark & \xmark & \cmark & \xmark & \xmark & \xmark & \xmark & \xmark & \xmark & \xmark & \cmark & Auto \\
Hamster       & 136k  & --  & \cmark & \cmark & \cmark & \xmark & \xmark & \xmark & \xmark & \xmark & \xmark & \xmark & \cmark & Auto \\
RH20T-P       & 38k   & 7   & \xmark & \cmark & \xmark & \xmark & \xmark & \cmark & \xmark & \cmark & \xmark & \xmark & \xmark & Human+Auto  \\
ECoT          & 60k   & 12  & \cmark & \xmark & \cmark & \cmark & \xmark & \cmark & \xmark & \xmark & \cmark & \cmark & \xmark & Auto   \\
AgiBot-World  & 1M    & 106 & \cmark & \xmark & \cmark & \xmark & \xmark & \cmark & \xmark & \xmark & \xmark & \xmark & \xmark & Human   \\
VLA-OS        & 10k   & --  & \cmark & \xmark & \cmark & \cmark & \xmark & \cmark & \cmark & \cmark & \cmark & \cmark & \cmark & Auto \\
ShareRobot    & 51k   & 102 & \xmark & \cmark & \xmark & \xmark & \xmark & \cmark & \cmark & \cmark & \xmark & \xmark & \cmark & Human+Auto \\
Robo2VLM      & 176k  & 463 & \xmark & \cmark & \xmark & \xmark & \xmark & \cmark & \xmark & \cmark & \xmark & \xmark & \cmark & Auto  \\
VeBrain       & 12k   & --  & \cmark & \cmark & \xmark & \xmark & \xmark & \cmark & \xmark & \cmark & \xmark & \xmark & \xmark & Human  \\
\midrule \rowcolor{gray!15}
\textbf{Ours} & \textbf{230k} & \textbf{571} & \cmark & \cmark & \cmark & \cmark & \cmark & \cmark & \cmark & \cmark & \cmark & \cmark & \cmark & \textbf{Human+Auto} \\
\bottomrule
\end{tabular}
\end{adjustbox}
\end{table*}

%% file: section/2_related_work.tex
\section{Related Works}
\label{sec_relatedworks}

\subsection{Embodied intermediate representations and datasets.} Embodied intermediate representations have emerged as an important interface between high-level planning and low-level action execution. Existing studies have explored a wide range of such representations, including 2D visual traces~\citep{rt-trajectory}, optical flow~\citep{im2flow2act}, object grounding and affordance box~\citep{sundaresan2023kite,huang2025roboground,huang2024manipvqa}, subtasks planning~\citep{zhang2024sprint,rth}, task-oriented pointing ~\citep{Point-It-Out,li2025hamster}, future images or goal states~\citep{zhao2025cot,lv2025f1,ma2026internvla,xu2026futurevla}, 3D reconstruction~\citep{yang2026robo3r}, and language-based reasoning chains~\citep{ji2025robobrain,emma-x}. These representations provide more structured and physically meaningful supervision than raw action tokens alone, and therefore serve as an intermediate abstraction for decomposing manipulation tasks and guiding action generation. Despite the strong perception ability of vision foundation models~\citep{sam2,oquab2023dinov2,morimitsu2025dpflow}, directly applying them to embodied manipulation remains challenging. General models can estimate certain visual cues, but their reliability often fluctuates across diverse physical scenes due to the domain gap between web-scale data and embodied observations. Consequently, recent works~\citep{robobrain2.0,robobrain2.5,rynnbrain,vebrain} has shifted from using off-the-shelf vision models alone toward curating task-specific embodied datasets and training VLMs to understand or generate embodied intermediate representations. For example, RoboBrain~\citep{robobrain2.0} emphasizes planning, affordance perception, and trajectory prediction as core capabilities.

Large-scale robotic manipulation datasets~\citep{rh20t,khazatsky2024droid,open_x_embodiment,gao2025genmanip,robomind2.0} provide diverse embodiments, scenes, and manipulation skills, forming the data foundation for robotic manipulation systems. However, these datasets are usually collected for imitation learning and typically lack labels for intermediate representations. This absence makes it difficult to train VLMs to explicitly bridge abstract instructions and concrete physical behaviors. To compensate for this limitation, a line of work augments existing robot demonstrations with additional annotations. RH20T-P~\citep{rh20tp} and RT-H~\citep{rth} provided extra primitive-level subtasks or motion-level language descriptions, providing temporally structured supervision for hierarchical planning. LLaRVA~\citep{niu2024llarva}, Hamster~\citep{li2025hamster}, and related trajectory-centric methods extract 2D gripper traces from demonstrations for VLM pretraining, enabling models to predict spatially grounded action traces. ECoT~\citep{ecot} and Emma-X~\citep{sun2024emma} further combine trajectory clips with grounded chain-of-thought and spatial reasoning. Other works focus on a single type of representation or high-level embodied understanding: RoboPoint~\citep{yuan2024robopoint} and Point-It-Out~\citep{Point-It-Out} study pointing or visual trace generation for embodied grounding, ManipVQA~\citep{huang2024manipvqa} injects manipulation-centric affordance and physical concept knowledge into MLLMs, and RoboAnnotatorX~\citep{RoboAnnotatorX} formulates robotic scene understanding, long-horizon reasoning, and interaction-phase recognition as VQA-style supervision.

However, relying on single-modality representations or post-hoc extraction limits the richness of embodied planning. In contrast, RoboInter~\citep{li2026robointer} and RoboInter1.5 introduce \textit{\textbf{per-frame}} dense annotation data across diverse intermediate representations, covering both low-level spatial cues and high-level task planning within a unified suite. This design advances both comprehensive embodied understanding and end-to-end action learning, and offers a practical foundation for bridging high-level planning with low-level execution in generalizable robotic manipulation.

\subsection{Embodied reasoning and world modeling for actions.} To bridge embodied reasoning and low-level action execution, recent research has focused on two parallel paradigms. The first paradigm leverages the profound semantic reasoning capabilities of VLMs in VLA models, which are generally divided into implicit and explicit reasoning. Implicit methods operate as black boxes~\citep{pi_0,roboflamingo,li2025cronusvla,11560913}, fine-tuning pretrained VLMs across various datasets to directly regress raw actions. In contrast, explicit methods prioritize interpretability, as seen in $\pi_{0.5}$~\citep{pi_0.5} and Rekep~\citep{rekep}. Building on this, models like ECoT~\citep{ecot} and VLA-OS~\citep{gao2025vlaos} introduce explicit text-based Chain-of-Thought (CoT), prompting VLMs to autoregressively generate step-by-step textual planning before discrete actions. While this language-driven planning inherits strong zero-shot generalization of VLMs, it suffers from a critical lack of perceptual priors, as language instruction is an inherently sparse representation~\citep{psiris2026foundation} and struggles to convey dense, high-frequency physical cues. Consequently, purely language-driven CoT is difficult to ground in physical reality, necessitating auxiliary perceptual structures to capture fine-grained visual information.

Another paradigm advocates for generative world modeling to act as an internal predictive simulator~\citep{gao2026dreamdojo, pai2025mimic, li2025comprehensive, team2025gigaworld, lou2026dream}. These models attempt to explicitly simulate the physical evolution of the environment. Architectures like Cosmos-Policy~\citep{kim2026cosmos}, Genie Envisioner (GE)~\citep{liao2025genie}, lingbot-va~\citep{lingbotva}, $\pi_{0.7}$~\citep{pi07} and DreamZero~\citep{dreamzero} jointly predict visual futures and optimal actions, while platforms like GigaWorld-0~\citep{team2025gigaworld} and Gaussian World Models~\citep{lu2025gwm} tackle multi-view coherence and spatial rigidity through 3D representations and physical foresight rewards. Despite offering a visually grounded reasoning mechanism, generative approaches suffer from systemic limitations when extended to long-horizon rollouts, largely due to the inadequacy of their conditioning signals. Language-conditioned video generation often hallucinates intermediate physical dynamics because text provides only coarse spatial constraints, while action-conditioned simulators rely primarily on raw motor actions, which lack intrinsic geometric semantics and impose an unintuitive representation on continuous dynamics~\citep{psiris2026foundation}. As a result, errors accumulate over time, ultimately making the generated futures unreliable for precise control.

RoboInter1.5 goes beyond CoT planning by systematically investigating the role of intermediate representations in embodied reasoning, particularly for complex, long-horizon tasks in diverse real-world settings. In addition, we study world modeling based on intermediate representations, which effectively captures and leverages task-relevant visual cues to enable accurate long-horizon rollouts and stronger generalization capabilities.

%% file: section/3_dataset.tex
\section{Dataset}
\label{sec_data}
\begin{figure*}[t]
    \centering
    \includegraphics[width=0.98\linewidth]{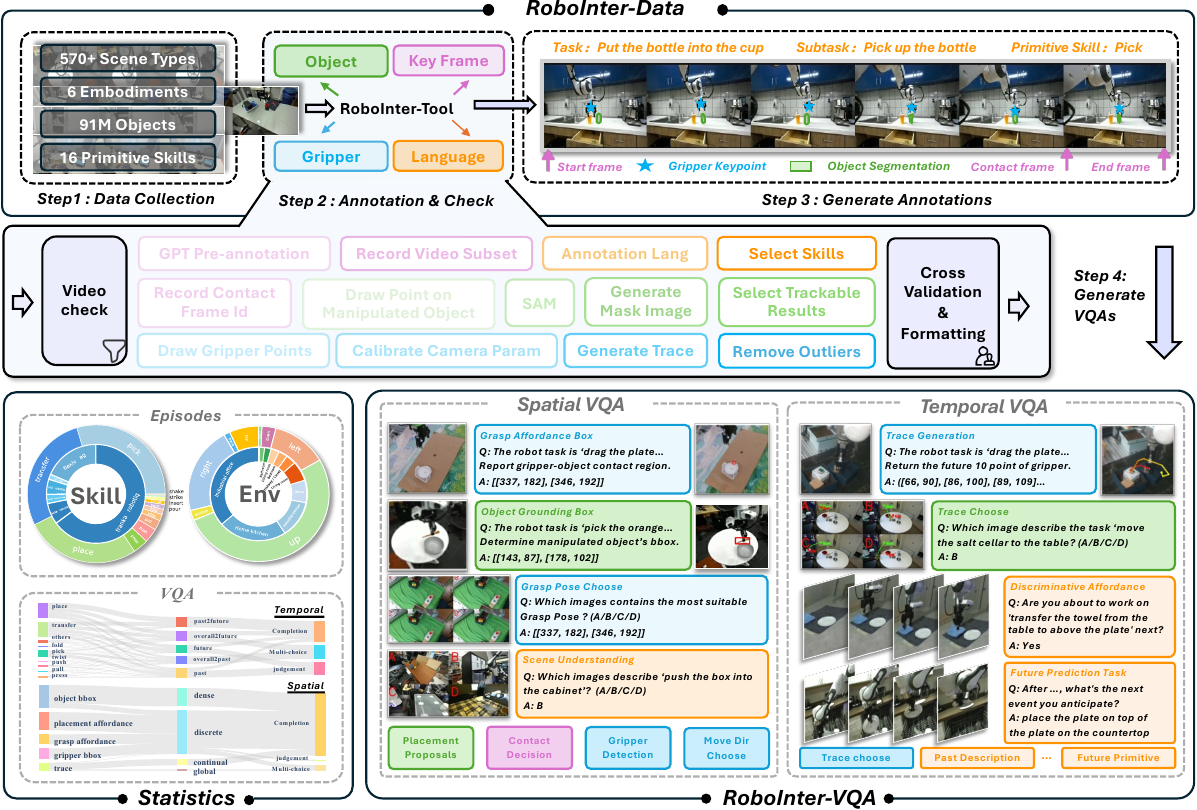}
    \caption{\textbf{Overview of RoboInter-Data and RoboInter-VQA.} We annotate 230k manipulation episodes with 10 types of intermediate representation annotations through Data Collection and Annotation \& Check. A large-scale, diverse set of VQA spanning spatial and temporal dimensions is further constructed. Statistics of raw episodes and curated VQA are also provided.}
    \label{fig:dataset}
\end{figure*}
\subsection{RoboInter-Data}
As illustrated in Figure~\ref{fig:dataset}, RoboInter-Data builds upon extensive manipulation datasets and provides large-scale, high-quality annotations of diverse intermediate representations.

\noindent\textbf{Data collection.} To enhance dataset diversity, we collected two types of raw manipulation data: (1) \textit{\textbf{In-the-Wild setting}} (i.e., diverse indoor scenarios), emphasizing the diversity of scenes and instructions, mainly collected from Droid~\citep{khazatsky2024droid}, RH20T~\citep{rh20t}, and OXE~\citep{open_x_embodiment}. (2) \textit{\textbf{Table-Top setting}} (i.e., tabletop interaction scenarios), highlighting the high quality and skills diversity, collected from RH20T~\citep{rh20t}. By integrating raw teleoperated video recordings of these datasets, followed by rigorous screening and pre-processing, we constructed a high-quality, large-scale database consisting of 230k manipulation episodes (third-person videos).

\noindent\textbf{Annotations \& Check with RoboInter-Tool.} For accurate and comprehensive labels, \textit{RoboInter-Tool} is employed to perform the following annotations: \textit{\textbf{(1) Task decomposition \& key-frame annotation.}} Each manipulation video is decomposed into clips using 15 predefined primitive skills, and ChatGPT~\citep{GPT-4} is employed to produce preliminary references for language annotations. Human annotators utilize \textit{RoboInter-Tool} to segment video clips, assigning each clip to a primitive skill, and simultaneously completing clip-level and video-level language annotations. The \textit{contact frame} where the robot arm contacts the manipulated object is also recorded. \textit{\textbf{(2) Recognizing the manipulated object.}} After recording the object that the robot arm interacts with, \textit{RoboInter-Tool} automatically transports the annotation to \textit{SAM2}~\citep{sam2} for object segmentation and tracking, and the result is asynchronously returned for review. A re-annotation and inspection mechanism reduces the impact of segmentation or tracking errors on the quality. 
\textit{\textbf{(3) Locating the end-effector.}} As many raw recordings lack reliable camera parameters, directly projecting 3D coordinates to obtain accurate 2D end-effector traces is often infeasible. We estimate a calibration matrix to improve projection accuracy and complement parameter-missing episodes via gripper detection and point tracking, enabling reliable reconstruction of the 2D trace. Details are provided in Appendix.

\noindent\textbf{Post-processed annotations.} By reorganizing the above annotations, we derive additional intermediate representations: \textit{\textbf{(1) Grasp annotation.}} The grasp affordance box is inferred from the 2D end-effector location at the annotated contact frame. The contact points are the pre-defined key points of the gripper at the moment of contacting, while the corresponding robot state (i.e., the 6D end-effector pose) defines the grasp pose. \textit{\textbf{(2) Placement annotation.}} The object position at the end of the subtask is treated as the target placement location. \textit{\textbf{(3) Gripper annotation.}} Anchor points enclosing the gripper are identified and projected from 3D to 2D using camera parameters and robot states, forming the gripper bounding box on the 2D image coordinates.

\noindent\textbf{Statistics of RoboInter-Data.} As shown in Figure.\ref{fig:dataset}, we provide high-quality, scene-diverse intermediate representation annotations for 230k manipulation episodes. This dataset includes 6 types of robot arms, 571 types of scenes, and 15 types of primitive skills. With the assistance of the RoboInter-Tool, we produce nearly 61M-frame object grounding annotations, about 70M-frame gripper trace annotations, 190k affordance boxes and placement proposals, and nearly 760k language clip annotations, providing both large scale and high quality.

\subsection{RoboInter-VQA}

\noindent\textbf{VQA task construction.} As illustrated in the \textit{RoboInter-VQA} section of Figure.\ref{fig:dataset}, we convert the annotations into diverse VQA tasks to enhance VLM capabilities. Tasks are organized along two axes: (i) intermediate representation type (spatial vs. temporal) and (ii) target capability (understanding vs. generation).
\textit{\textbf{(1) Spatial VQA for understanding.}} We design three selection tasks and one judgment task to train spatial comprehension, including selecting the correct object bounding box or grasp pose, matching scenes to instructions, and determining whether contact occurs.
\textit{\textbf{(2) Spatial VQA for generation.}} It includes five prediction tasks that require generating spatial intermediate representations for downstream execution and complete specific content based on spatial reasoning, which includes object bounding box, grasp pose, placement proposal, key points, and the gripper bounding box.
\textit{\textbf{(3) Temporal VQA for understanding.}} To evaluate how the VLMs understand motion traces and the relationships between subtasks and observations, we design selection tasks and judgment tasks for temporal information. We design five selection tasks for movement directions of grippers, the matching of trace and description, subtask/primitive discrimination, and execution stage identification. Four judgment tasks assessing task success and next-step feasibility.
\textit{\textbf{(4) Temporal VQA for generation.}} We formulate tasks that require trace generation and multi-step planning under varying levels of contextual completeness. Prompts condition on different amounts of prior information (e.g., past subtasks or overall instructions) and ask the model to predict the subsequent steps or multi-step planning. Video-based visual inputs are also used to summarize past events and predict feasible next steps. Trace generation is evaluated under both easy and challenging settings (with or without initial waypoints). Details are included in the Appendix.

\vspace{3pt}
\noindent\textbf{Statistics of RoboInter-VQA.} Our VQA data is also large in scale, comprising approximately 1M spatial generation entries, 172k spatial understanding entries, 131k temporal generation entries, and 935k temporal understanding entries. To prevent information leakage between training and validation, we carefully designate 7,246 videos as the evaluation pool with the remaining data used for training, and sample validation sets for each question category from this pool.

\subsection{RoboInter-CV}
\noindent\textbf{Data construction.} Currently, long-horizon robotic world imagination remains difficult when models are conditioned only on natural language or actions. Language instructions are usually too coarse to describe detailed object motion, contact timing, and spatial constraints, while robot actions are low-level embodiment-specific signals whose visual consequences are difficult to align with image-space dynamics. This mismatch becomes more severe over long horizons, where small temporal or spatial errors accumulate and lead to object drift, inconsistent contacts, or implausible motion. 
To provide a more direct and visually aligned control signal, we construct \textbf{RoboInter-CV}, a control-video dataset derived from \textit{RoboInter-Data}. For each manipulation episode, we first sample representative object points from the annotated segmentation masks and track them across time, converting dense object masks into sparse object-centric motion cues. We then render these tracked object points together with gripper traces into a compact video on a black canvas, using the same temporal axis as the original trace annotations. To ensure data quality, we filter training clips by checking whether the corresponding future window contains sufficient valid visual controls. 
Specifically, we build frame-level validity masks from segmentation and trace annotations, where a valid frame requires a non-trivial object mask and a non-empty trace signal. We also discard episodes with missing RGB videos, missing annotations, overly short latent sequences, or no valid anchor frame. The resulting control videos remove appearance details while preserving object-centric spatial cues and gripper motion, offering dense visual conditions that bridge coarse language, low-level actions, and future observation prediction.

\noindent\textbf{Data Statistics}. It contains 65k clip-level samples from 16.9k unique manipulation episodes, covering both DROID and RH20T sources. Each retained sample is paired with a rendered control video, gripper trace, robot action sequence, language instruction, subtask, RGB observation, and future latent chunk, ensuring complete alignment among visual controls, low-level actions, and future states. We render the control videos at the same resolution as that used for world-model training, and employ them as structured visual conditions for controllable future prediction in \textit{RoboInter-World}.

%% file: section/4_method.tex
\section{Method}
\label{sec_method}

In this section, we present how intermediate representations are used by embodied models. We first introduce the shared VLM-based Planner, which learns to understand and generate intermediate representations
from \textit{RoboInter-VQA} supervision. We then describe two downstream instantiations: \textit{RoboInter-VLA}, which follows a \textit{plan-then-execute} paradigm for action generation, and \textit{RoboInter-World},
which performs intermediate-conditioned future observation generation with explicit visual controls.

\subsection{RoboInter-VLM Planner}
\label{sec_method_planner}

\noindent\textbf{VLMs as Planner.}
The Planner model acquires embodied capabilities through a visual question answering formulation with a co-training strategy. To capture both spatial and temporal information, we adopt VLM architectures that
support single- and multi-image inputs, including the Qwen-VL series~\citep{qwen2vl} and LLaVA-One-Vision~\citep{llavaov}. Each model consists of a base LLM, a vision encoder, and an MLP-based vision--language
projector. The Planner generates outputs autoregressively and is optimized using a cross-entropy loss.

The Planner serves as a shared source of intermediate representations for downstream modules. Since \textit{RoboInter-VLA} inherits a VLM backbone from the Planner, it can incorporate intermediate representations through implicit feature transfer, explicit joint reasoning, or modular planner-to-executor conditioning, as detailed in Section~\ref{sec_method_vla}. In contrast, RoboInter-World mainly focuses on video generation and does not adopt the same implicit or end-to-end integration with VLM used in \textit{RoboInter-VLA}. Instead, it uses intermediate representations as explicit visual controls for future observation generation. Therefore, we focus on how these controls are injected into the world model and compare controls from ground-truth intermediate representation annotations and Planner-predicted intermediate representations.

\subsection{RoboInter-VLA: Plan-Then-Execute}
\label{sec_method_vla}

\begin{figure*}
  \centering
  \includegraphics[width=0.99\linewidth]{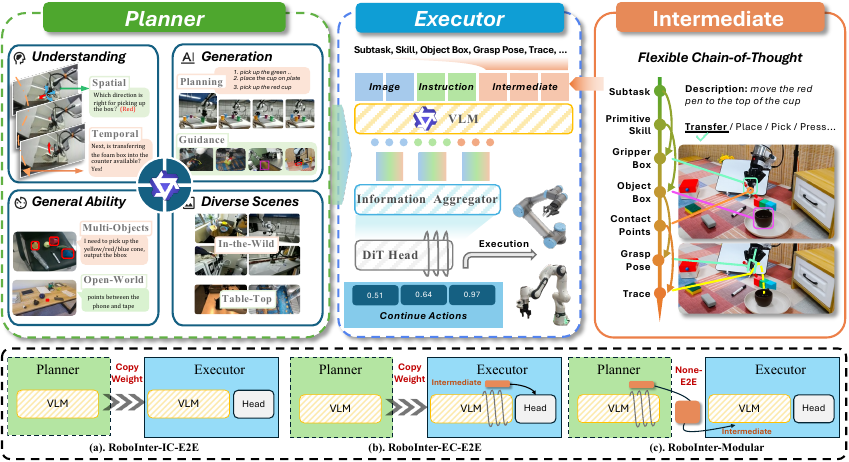}
  \caption{\textbf{Framework of RoboInter-VLA.} Our model follows a \textit{\textbf{plan-then-execute}} paradigm with a VLM-based \textit{Planner} and an \textit{Executor}. The Planner exhibits enhanced
understanding and generation for manipulation, strong general grounding abilities, and robust perception across diverse scenes. The Executor shares the VLM backbone with the Planner. Three variants are
supported, and intermediate representations in Flexible Chain-of-Thought (F-CoT) bridge planning and execution.}
  \label{fig:method}
\end{figure*}

As illustrated in Figure~\ref{fig:method}, \textit{RoboInter-VLA} is a family of models following a \textit{plan-then-execute} paradigm, consisting of a \textit{Planner} and an \textit{Executor}. Rather than a
monolithic design, \textit{RoboInter-VLA} supports multiple variants and enables flexible adaptation from planning to execution. The \textit{Planner} performs high-level decision-making by combining general and
embodied reasoning to produce intermediate representations, which guide the \textit{Executor} in translating multimodal observations and language instructions into low-level actions.

\noindent\textbf{Executor Architecture.} Our model designs mainly follow InternVLA-M1~\citep{chen2025internvla} and CogACT~\citep{cogact}. To systematically and lightweightly derive from the Planner, we build Executor on a Qwen2.5-VL backbone with a Diffusion Transformer (DiT) action head~\citep{DiT}. We further utilize an \textit{information aggregator} that gathers the hidden states of all input and output tokens, as well as intermediate representations, and compresses them into conditioning features with a controllable length. The \textit{Executor} consumes multi-view visual observations (e.g., primary and wrist), language instructions, and intermediate representations (based on primary observation), and produces multi-step action chunks via a diffusion loss.

\noindent\textbf{Plan-Then-Execute Paradigms.} As shown in Figure~\ref{fig:method}, by leveraging the pretrained Planner, we provide three paradigms to enhance downstream action execution:
(1)~\textit{RoboInter-IC-E2E (Implicitly-Conditioned End-to-End)}, which directly injects the VLM from a pretrained Planner into the end-to-end Executor, using it as a stronger vision-language feature
extractor. This approach can yield robust embodied perception and more accurate task-relevant visual cues.
(2)~\textit{RoboInter-EC-E2E (Explicitly-Conditioned End-to-End)}, where the Executor is initialized with the VLM of the Planner and jointly optimizes both reasoning and action
generation.
(3)~\textit{RoboInter-Modular (Modular Planner-to-Executor)}, a non-E2E hierarchical design that treats the Planner and Executor as independent modules. During training, the Executor conditions on ground-truth
intermediate representations to generate actions, whereas during inference, it uses the predicted results of the Planner.

\noindent\textbf{Flexible chain-of-thought for intermediate representations.} To support the explicitly conditioned and modular architectures, we introduce \textbf{F-CoT}, a chain-of-thought composed of multiple intermediate representations. F-CoT plays two roles: (i) as VQA supervision for training the Planner, and (ii) as action-aligned guidance for the Executor. In \textit{RoboInter-IC-E2E}, the VLM generates F-CoT content, which is directly consumed by the DiT head. In \textit{RoboInter-Modular}, the Planner produces the F-CoT content and the Executor conditions on it. F-CoT flexibly combines representations such as subtasks, skills, object bounding boxes, affordance boxes, motion traces, etc., in textual or visual form, allowing users to select subsets tailored to specific embodied tasks. We denote textual F-CoT as \textit{RoboInter-Te-Modular} and visual-prompted F-CoT as \textit{RoboInter-Im-Modular}.

\subsection{RoboInter-World: Controlled Imagination}
\label{sec_method_world}
\textit{RoboInter-World} extends \textit{RoboInter} from action execution to future observation generation. Given observed history frames and intermediate controls, \textit{RoboInter-World} generates future observations in the latent video space. Let \(z^{\mathrm{hist}}\) denote the observed history latents and \(x_t\) denote the noisy future latent at diffusion timestep \(t\). The denoising model is formulated as:
\begin{equation}
f_\theta(x_t, t \mid z^{\mathrm{hist}}, c, a, u),
\end{equation}
where \(c\) is the language instruction, \(a\) is the robot action sequence when used, and \(u\) is the control video rendered from intermediate representations. After iterative denoising, the predicted future latents are decoded by the 3D VAE into RGB observations.
The model is optimized by
\begin{equation}
  \mathcal{L}_{\mathrm{world}}
  =
  \mathbb{E}_{y,\epsilon,t}
  \left[
  \left\|
  v_t -
  f_\theta(x_t, t \mid z^{\mathrm{hist}}, c, a, u)
  \right\|_2^2
  \right],
\end{equation}
where \(v_t\) is the flow-matching target from the noisy future latent \(x_t\) to the clean future latent \(y\).

\begin{figure}
  \centering
  \includegraphics[width=0.98\linewidth]{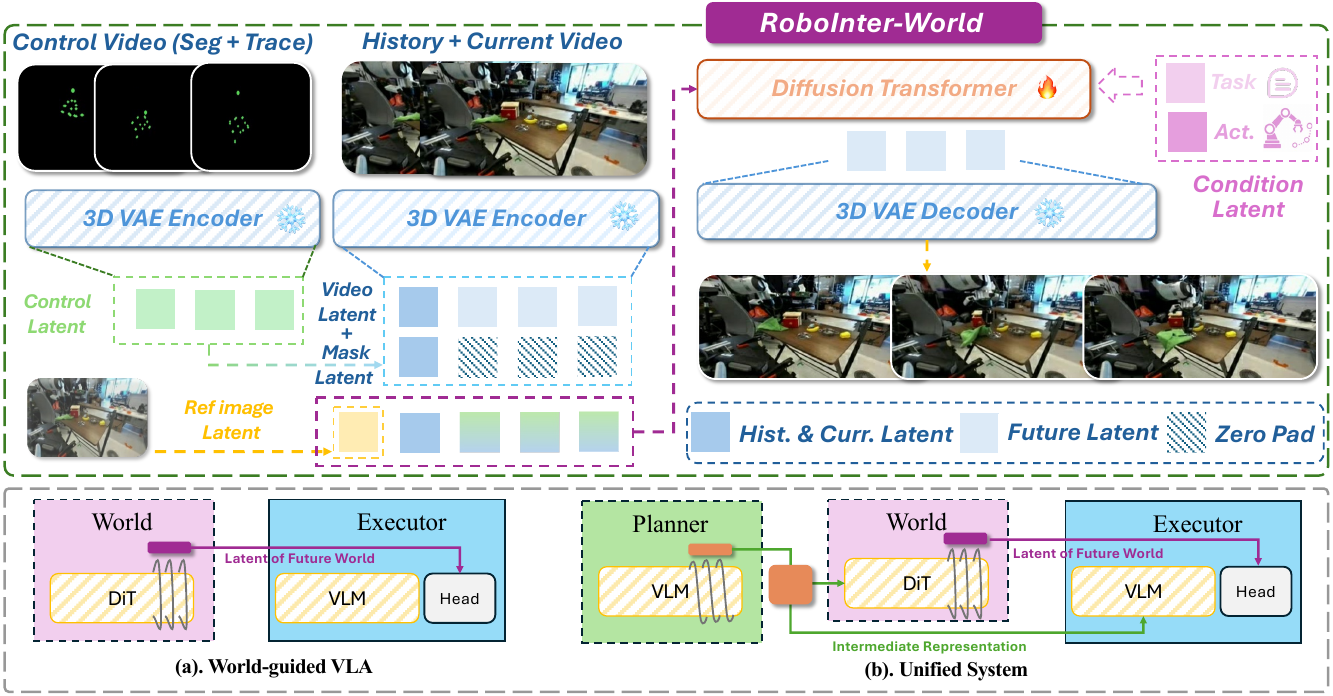}
  \caption{\textbf{Overview of RoboInter-World}. The mask latent is employed to encode temporal information, explicitly distinguishing between historical/current states and future contents. The control video is encoded into control latents, which are subsequently concatenated with the input latents along the channel dimension. Concurrently, conditioning latents (when available), comprising language instructions and robot actions, guide the diffusion transformer via the cross-attention blocks in the DiT architecture. Two variances in which \textit{RoboInter-World} guides the action executor are also illustrated.}
  \label{fig:wm}
\end{figure}

\vspace{3pt}
\noindent\textbf{Control Video from Intermediate Representations.} The control video is constructed using the intermediate representations from \textit{RoboInter-Data}. As described in RoboInter-CV, for object-centric control, we translate dense visual information into concise positional cues by uniformly sampling points within object bounding boxes or segmentation masks, and for gripper-centric control, we leverage 2D gripper traces. Crucially, the formulation of these conditions significantly impacts model performance. We empirically observe that directly injecting raw numerical coordinates via MLPs struggles to provide sufficient spatial grounding for the diffusion model. Furthermore, visual feature-replacement techniques (e.g., Wan-Move~\citep{chu2026wan}) tend to introduce noisy conditioning when dealing with objects undergoing substantial topological or state changes, such as cloth folding or gripper articulation. To overcome these limitations, we explicitly render both the extracted object points and the gripper traces onto a blank canvas, temporally aligned with the original observations. By encoding intermediate representations in a visual form, we can effectively exploit the strong spatial priors embedded in pretrained vision models, thereby accelerating training convergence while avoiding the noise artifacts introduced by direct feature replacement.

\vspace{3pt}
\noindent\textbf{World Model Architecture.} We instantiate \textit{RoboInter-World} based on a latent video diffusion transformer. As shown in Figure~\ref{fig:wm}, the model processes heterogeneous inputs through modality-specific pathways. Language instructions $c$ and action sequences $a$ are encoded and concatenated to form the cross-attention context:
$$h^{\mathrm{ctx}} = \left[ \phi_{\mathrm{text}}(c); \phi_{\mathrm{act}}(a) \right].$$
The VAE-encoded control latent $s^{\mathrm{ctrl}}$ and a mask latent $m$, which explicitly distinguishes historical states from future contents, are concatenated with the video latents along the channel dimension. The resulting features are then flattened into spatio-temporal tokens. Specifically, the token sequences of the noisy future $x_t$ and the clean history $z^{\mathrm{hist}}$ are concatenated along the sequence dimension, followed by the integration of CLIP-based reference image tokens $z^{\mathrm{ref}}$. Finally, taking these integrated spatio-temporal tokens as input and conditioned on $h^{\mathrm{ctx}}$, the DiT module predicts the flow target $\hat{v}_t$.

\vspace{3pt}
\noindent\textbf{Control Protocol.} During training, \textit{RoboInter-World} follows a teacher-forcing setup. However, to mitigate the discrepancy between training and testing distributions, we introduce stochastic perturbations to the intermediate representations. This strategy enhances the model's robustness against imperfect, planner-generated controls during inference.
So during inference, we evaluate two control protocols. In the oracle-control protocol, control videos are rendered from ground-truth annotations, measuring the upper-bound utility of high-quality intermediate representations. In the planner-control protocol, the RoboInter Planner generates future intermediate representations, such as future trace points or bounding boxes of future key frames, which are post-processed and rendered into control videos with the same procedure. Furthermore, we explore how world models can guide VLA models. Unlike $\pi_{0.7}$~\citep{pi07}, which treats subgoal images predicted by a world model as metadata for the VLM, we directly feed the image latents predicted by the world model into the executor’s action head. These latents function as image-level intermediate representations that interact with action prediction more directly.

%% file: section/5_exp.tex
\section{Benchmarking and Experiments}
\label{sec_exp}

\input{tables/main_general_benchmark}

\subsection{Benchmarking the Planner}
\label{subsec_exp4planner}
\noindent\textbf{Enhanced grounding and embodied capability.} As shown in Table~\ref{tab:multi_task_performance_reordered}, we evaluated on third-party spatial reasoning benchmarks, including Where2Place~\citep{yuan2024robopoint} and RoboRefIt~\citep{lu2023vl} (spatial point and grounding reasoning) and RoboVQA~\citep{robovqa} (temporal task planning). Across all three benchmarks, our models substantially outperformed the base models (Qwen2.5-VL-3B/7B~\citep{qwen2vl} and LLaVA-OneVision-7B~\citep{llavaov}). Notably, RoboBrain2.0~\citep{robobrain2.0} is also an embodied VLM (i.e., Planner). At the 3B scale, \textit{RoboInter-Qwen-3B} achieved a 49.1\% improvement over RoboBrain2.0 on RoboRefIt and a 12.7\% improvement on RoboVQA. At the 7B scale, the corresponding gains reached 76.8\% and 42.8\%, respectively. For grounding, all three RoboInterVLM variants exceeded their respective base models on Refcoco~\citep{coco}. Particularly, \textit{RoboInter-Qwen-7B} eventually ranked second overall, with a 27.4\% relative improvement over RoboBrain2.0-7B. On general benchmarks, our models remained relatively stable on most benchmarks, indicating that \textit{\textbf{our curated VQA data enhances the abilities of embodied reasoning and grounding}}, meanwhile, the general capabilities of our VLMs are slightly affected.

 \vspace{3pt}
\noindent\textbf{RoboInter-VQA benchmark at the spatial and temporal level.} As shown in Table~\ref{tab:robointer_spatial_temporal_revised}, for spatial-based generation tasks, closed-source and general VLMs without embodied experience rarely produce accurate intermediates (typically below 40\%), underscoring the importance of additional intermediate representation annotations. On simpler questions, Gemini-2.5-Flash and RoboBrain-2.0-7B lead on \textit{Grasp Pose (choice)} with 32.7\% and 23.3\% ACC. For \textit{Grounding Choice}, LLaVA-OV-7B achieves 31.9\%, while Gemini-2.5-Flash is strongest at 69.4\%; most other models remain near random choosing (25\%) given limited understanding of manipulation scenes. For temporal, closed-source API and general VLMs largely fail to generate future traces or task planning; RoboBrain-2.0 attains a much better DTW in \textit{Trace Generation} than Qwen-VL-2.5 (541 v.s. 1702). On \textit{Visual Trace Choice}, Gemini-2.5-Flash remains competitive (49.4\%). For \textit{Planning Choice} and \textit{T/F of Task Planning}, as planning aligns closely with general LLM abilities, most models transfer common-sense knowledge and show better performance. Overall, \textit{\textbf{current closed-source and general VLMs typically lack enough embodied abilities}}. Curated from diverse annotations, \textbf{\textit{RoboInter-VQA markedly improves the VLM abilities of understanding and generating intermediate representations}}.

\noindent\textbf{Qualitative results.} As illustrated in Figure.\ref{fig:vlm}, we provide qualitative examples demonstrating the capability of RoboInter-VLM to perform motion planning and subtask planning in in-the-wild scenarios. These examples highlight its spatial understanding of manipulation and its temporal reasoning ability for generating coherent task plans. Benefiting from training on both general multimodal data and RoboInter-VQA, RoboInter-VLM exhibits strong embodied understanding and generation capabilities across temporal and spatial dimensions.

\input{tables/main_RoboInterBenchmark}

\subsection{Open-Loop Evaluation of the Executor}
\label{subsec_exp4excutor}

\input{tables/tab_OpenL_OH}

\noindent\textbf{Experimental settings}. In this section, we examine how a pretrained Planner improves the Executor and compare different VLA paradigms. As our annotated corpus spans more than 500 distinct scenarios, comprehensive real-world validation across all scenarios is infeasible, as emphasized by HPT~\citep{hpt}. Following the discrete token-accuracy evaluation in OpenVLA~\citep{openvla}, we utilize an \textit{\textbf{Open-Loop Score (OLS)}} to evaluate the generation of \textit{continuous action chunks}, in which per-step actions are assessed independently and compared with the ground-truth actions. OLS is computed as the average value over 100K transitions from evaluation videos, ensuring statistical stability. More details in the Appendix.
Nine Executor variants are evaluated: (a).\textit{Vanilla}: omits any pretrained VLM from Planner, performing action learning only; (b-e).\textit{RoboInter-IC-E2E}, \textit{EC-E2E}, \textit{Te-Modular} and \textit{Im-Modular} are stated in Section.\ref{sec_method_vla}; (f).\textit{Oracle+Executor}: not end-to-end, both training and inference are guided by GT intermediate representations; (g).\textit{QwenVL+Executor}: training is GT-guided, and inference employs intermediates from original Qwen2.5VL; \modify{(h).\textit{VLA-OS}: we train VLA-OS~\citep{gao2025vlaos} in our setting and use it as an additional baseline}; (i).\textit{Oracle+VLA-OS}: not end-to-end, VLA-OS are guided by GT intermediate representations. Two open-loop evaluation settings: (1) \textit{In-the-Wild}: focus on scene and object generalization, we compare the convergence performance under identical training steps. (2) \textit{Table-Top}: focus on the tabletop environment and cross-embodiment ability, we mainly examine the evaluation curve during training. The annotated corpus is divided into \textit{In-the-Wild} and \textit{Table-Top} subsets. We sample approximately 10\% of episodes from each subset for Executor training (25k in total), of which 8\% are reserved for evaluation.

\noindent\textbf{Planner consistently improves the Executor’s action generation}. The \textit{In-the-Wild} setting is shown in Table.\ref{tab:OpenL_OH}. The \textit{Vanilla} achieves a lower mOLS score than IC-E2E (0.3086 v.s. 0.3218), which incorporates intermediate representations, indicating that \textit{\textbf{pretrained VLM Planner can enhance the learning capability of the VLA Executor}}. The mOLS score of \textit{EC-E2E} is higher than \textit{IC-E2E} (0.3340 v.s. 0.3218), showing that the \textit{\textbf{explicit intermediate representations are more helpful for action guidance than the implicit}}. For \textit{Oracle+Executor}, the non-E2E architectures, utilizing ground-truth annotation in the Executor, achieve substantially the highest scores. This indicates that \textit{\textbf{our annotations are informative and stable}}. \textit{Te-Modular} surpasses \textit{EC-E2E} (0.3543 v.s. 0.3340), implying that \textit{\textbf{decoupling planning and execution facilitates dedicated optimization}} of each capability and mitigates mode conflict. \textit{Im-Modular} performs slightly worse than \textit{Te-Modular} (0.3439 v.s. 0.3543), as visual prompting embeds within information-dense images, diluting their relative contribution. \textit{QwenVL+Executor} employs Qwen2.5-VL as a zero-shot Planner, and its overall performance is not comparable with other Non-E2E models, \textit{\textbf{showing that the embodied reasoning ability of our Planner is better than the general VLMs}}. The evaluation curves of the \textit{Table-Top} are shown in Figure.\ref{fig:OpenL_TT}, as Table-Top scenes are easier to interpret, most methods eventually achieve high scores. \textit{RoboInter-Te-Modular} and \textit{Oracle+Executor} converge faster and reach higher performance. \textit{EC-E2E} converges more slowly but ultimately approaches the performance of \textit{Te-Modular}. \textit{IC-E2E} shows stronger early-stage results and maintains a consistent advantage over \textit{Vanilla} after 20k steps.

\begin{figure}[t]
  \centering
  \begin{minipage}[t]{0.49\linewidth}
    \vspace{0pt}
    \centering
    \captionof{table}{\modify{\textbf{Ablation of intermediate representation}. 
    We report OLS under multiple thresholds. Six representations are evaluated, where finer-grained categories yield larger gains. In the table below, we set abbreviations: S., Subtask; P., Primitive Skill; O.B., Object Box; G.B., Gripper Box; Aff., Affordance; and Tr., Trace.}}
    \label{tab:ablation_intermidate_comp}
    \renewcommand{\arraystretch}{1.15}
    \setlength{\tabcolsep}{1pt}
    \scriptsize
    \begin{tabularx}{\linewidth}{l|cccc}
      \toprule
      \textbf{Variant} & \multicolumn{4}{c}{\textbf{OLS}}\\
      \cmidrule(lr){2-5}
                       & @0.1 & @0.05 & @0.03 & @0.01\\
      \midrule
      Vanilla & 0.6793 & 0.3608 & 0.1753 & 0.0189 \\\rowcolor{gray!15}
      + S. & 0.6965 & 0.3676 & 0.1770 & 0.0171 \\
      + S. + P. & 0.6983 & 0.3681 & 0.1779 & 0.0194 \\\rowcolor{gray!15}
      + S. + P. + O. & 0.7025 & 0.3849 & 0.1988 & 0.0294 \\
      + S. + P. + O. + G.B. & 0.7212 & 0.4032 & 0.2048 & 0.0272\\\rowcolor{gray!15}
      + S. + P. + O. + G.B. + Aff. & 0.7245 & 0.4083 & 0.2114 & 0.0297 \\
      + S. + P. + O. + G.B. + Aff. + Tr. & \textbf{0.7511} & \textbf{0.4640} & \textbf{0.2705} & \textbf{0.0587}\\
      \bottomrule
    \end{tabularx}
  \end{minipage}
  \hfill
    \begin{minipage}[t]{0.5\linewidth}
    \vspace{1pt}
      \centering
      \includegraphics[width=\linewidth]{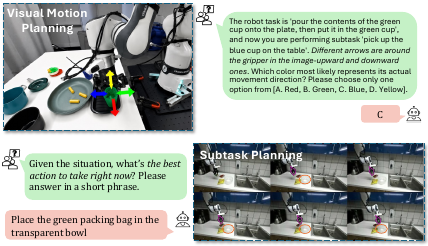}
      \caption{\textbf{Inference cases}. For the \textit{Visual Motion Planning}, RoboInter-VLM accurately infers the action direction of the gripper, demonstrating a strong understanding of spatial relations. For the \textit{Subtask Planning}, given a video input, RoboInter-VLM can reason about the underlying task intention.}
      \label{fig:vlm}
  \end{minipage}
  \vspace{-10pt}
\end{figure}

\noindent\textbf{Ablations on intermediate representations}. We ablate combinations of different intermediate representations within the open-loop Oracle+Executor setting. As shown in Table~\ref{tab:ablation_intermidate_comp}, coarse-grained representations such as \textit{Subtask} and \textit{Primitive Skill} provide only marginal improvements, as they offer stage-level guidance with limited actionable constraints during execution. In contrast, spatially grounded representations (\textit{Object Box}, \textit{Gripper Box}, and \textit{Affordance}) yield substantially larger gains by providing finer-grained cues. The most significant improvement comes from \textit{Trace}, which introduces dense, temporally grounded information and achieves the strongest overall performance. Additional results are provided in the Appendix.

\begin{figure*}
    \centering
    \includegraphics[width=0.98\linewidth]{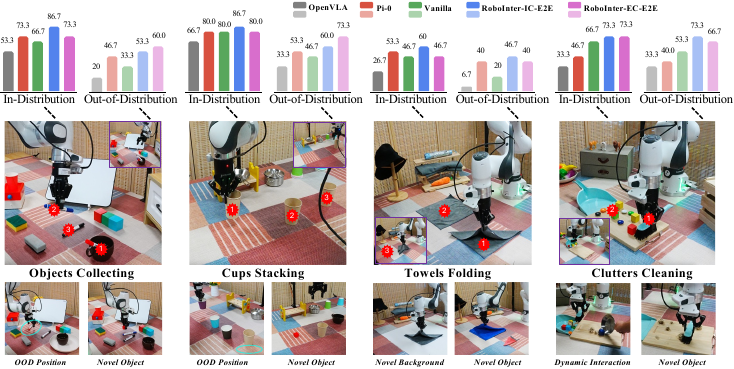}
    \caption{\textbf{Real-World experiments.} The top charts present results from 15 in-distribution (ID) and 15 out-of-distribution (OOD) trials. The bottom panel illustrates the OOD test setup. Notably, the performance drop from ID to OOD reflects each model’s generalization under distribution shift, where EC-E2E outperforms IC-E2E and exhibits a smaller ID→OOD degradation (8.3\% vs. 19.0\%), showing the consistent conclusion with the \textit{Open-Loop Evaluation}. Key steps are marked with a number, along with an end-execution thumbnail. Experiments of RoboInter-Modular are in the Appendix.}
    \label{fig:real world}
\end{figure*}

\subsection{Closed-Loop Real-World Evaluation of the Executor}
\label{subsec_exp4excutorRela}

\noindent\textbf{Experimental setting}. We study how our dataset and the pretrained Planner affect the closed-loop success rate. Experiments are conducted in a few-shot TableTop evaluation with a real-world Franka Research-3 arm. Observation input comprises a static third-person camera and a wrist-view camera, and no proprioceptive state. We focus on more practical E2E variants and evaluate five E2E models: (1) \textit{OpenVLA}~\citep{openvla}: Initialized from official pretrained weights and extended with an additional wrist-view input. (2) \textit{Pi-0}~\citep{pi_0}: Fine-tuned from the official checkpoints of Droid. (3) \textit{Vanilla}: our baseline. (4) \textit{RoboInter-IC-E2E}: Initialized from the Planner and further finetuned on in-distribution (ID) data. (5) \textit{RoboInter-EC-E2E}: Initialized from the Planner, and jointly optimizes action and CoT generation, with a 1:1 ratio of our annotated data and collected data. As shown in Figure.\ref{fig:real world}, we design four tasks: (a) Object Collecting: sequentially place three pens from a cluttered tabletop into a cup, out-of-distribution (OOD) tests are with novel objects, and alter spatial layouts. (b) Cup Stacking: Stack cups from left to right. OOD tests include novel objects, OOD positions, and continuous stacking. (c) Towel Folding: Fold two towels in sequence and stack them. OOD tests vary in the towel category and background. (d) Clutter Cleaning: Clean all items from the board with a brush. OOD tests introduce novel objects and disturbances. We provide more real-world
results of Tool Inserting and Object Sorting in the Appendix.

\noindent\textbf{Experimental results on ID and OOD testing.} 
Across all tasks, \textit{RoboInter-IC-E2E} consistently outperforms the \textit{Vanilla}. In ID evaluations, IC-E2E attains an average success rate of 77.3\%, compared with 65.0\% of Vanilla. Under OOD conditions, the gap widens, and IC-E2E achieves a 58.3\% success rate, while Vanilla reaches only 38.3\%, indicating the superior generalization of \textit{IC-E2E}. \textbf{\textit{The pretrained VLM from the Planner is pre-exposed to rich embodied data and therefore provides stronger perceptual priors.}} Although Pi-0, which is pretrained on Droid, also demonstrates solid ID and OOD performance, the \textit{IC-E2E} benefits from a broader representation training, thereby producing better overall results. \textit{EC-E2E} records a lower ID success rate than \textit{IC-E2E} (68.3\% vs. 77.3\%), which seems to be misaligned with the open-loop results in which \textit{EC-E2E} was superior. Actually, the open-loop protocol enforces strict decouple between training and validation, and therefore functions more like an OOD test. Correspondingly, under real-world OOD conditions, \textit{EC-E2E} exceeds \textit{IC-E2E} in Object Collecting (60.0\% vs. 53.3\%) and Cup Stacking (73.3\% vs. 60.0\%), and achieves a higher average success rate (60.0\% vs. 58.3\%). The ID-to-OOD drop is only 8.3\% for \textit{EC-E2E}, whereas \textit{IC-E2E} declines by 19\%. We attribute \textit{EC-E2E}’s weaker ID accuracy to the potential modality interference from the joint training of text generation and action prediction. \textit{\textbf{Diverse OOD knowledge from our dataset contributes to superior OOD robustness and generalization}}. We provide real-world results of \textit{RoboInterVLA-Modular} and more experiments within the WidowX platform in the Appendix.

\input{tables/wm_main}

\subsection{Benchmarking on world modeling}

\noindent\textbf{Main Results.} We evaluate the impact of intermediate representations across varying model scales (1.3B and 14B) and fine-tuning strategies (LoRA and Full Fine-Tuning), with comprehensive quantitative results summarized in Table~\ref{tab:main_results}.  We systematically investigate the oracle visual generation quality across various model architectures, fine-tuning strategies, and parameter scales. The \textit{Inter} control mode denotes the combined utilization of both 2D traces and segmentation masks. The relevant metrics are all standard evaluation criteria for video generation tasks. For more details, please refer to ~\cite{wan2025wan}.
As shown, relying solely on raw action conditioning severely bottlenecks generative quality. For instance, the 14B action-conditioned baseline (\textbf{RoboInter-W+Action}) yields an 18.26 PSNR and 0.171 LPIPS. Replacing sparse actions with our dense intermediate representations substantially boosts performance, improving PSNR to 21.05 and reducing LPIPS to 0.102 under the same 14B backbone.
Importantly, this structural advantage is orthogonal to model scale and tuning strategies. In the 1.3B regime, integrating \textit{Inter} consistently outperforms raw actions under both LoRA and Full Fine-Tuning (e.g., 20.00 vs. 19.08 PSNR in Full-FT). These results demonstrate that \textbf{\textit{intermediate representations provide robust dynamic constraints for accurate physical simulation.}}

\vspace{3pt}

\noindent\textbf{Impact of Context and Prediction Horizons.} In Table~\ref{tab:horizons}, extending prediction horizons reduces interaction frequency but increases learning difficulty and error accumulation. 
To investigate this trade-off, we ablate historical context and future prediction lengths using the 14B model. 
For short-term predictions with enough history (e.g., H4P3), the rich temporal prior provides sufficient regularity to extrapolate near-future dynamics, allowing the action-only baseline to perform adequately. However, when the historical context is limited to a single frame (H1P3), inferring temporal momentum becomes challenging. 
Under this heightened uncertainty, Seg+Trace provides essential geometric priors that \textbf{\textit{consistently improve generation fidelity, becoming critical for long-horizon predictions.}} 
As the prediction horizon extends, the action-only baseline suffers from severe compounding errors and structural degradation. 
In contrast, \textit{Seg+Trace} serves as a robust spatiotemporal scaffold that prevents drift, mitigating physical uncertainty and sustaining generation quality over extended sequences.

\noindent\textbf{Modality Ablation Study.} To isolate the specific contributions of individual intermediate modalities, we conduct a fine-grained ablation study across different model scales (Table~\ref{tab:ablation}). When applied independently, both segmentation masks (\textit{Seg}) and 2D traces (\textit{Trace}) significantly outperform the raw action-conditioned baselines. Mechanistically, these modalities prove highly synergistic: \textbf{\textit{segmentation masks enforce object-centric spatial constraints to prevent visual morphing, while 2D traces impose explicit temporal kinematics on the end-effector.}} By combining them (\textit{Seg+Trace}), RoboInter-W achieves the highest overall fidelity across all parameter scales. For instance, on the 14B model, the combined formulation reaches a peak PSNR of 21.05 and reduces LPIPS to 0.102. This confirms that a comprehensive spatiotemporal constraint can effectively eradicate the structural drift that typically plagues purely action-conditioned generators.
\input{tables/wm_abl_modality}

\begin{figure*}
  \centering
  \includegraphics[width=0.85\linewidth]{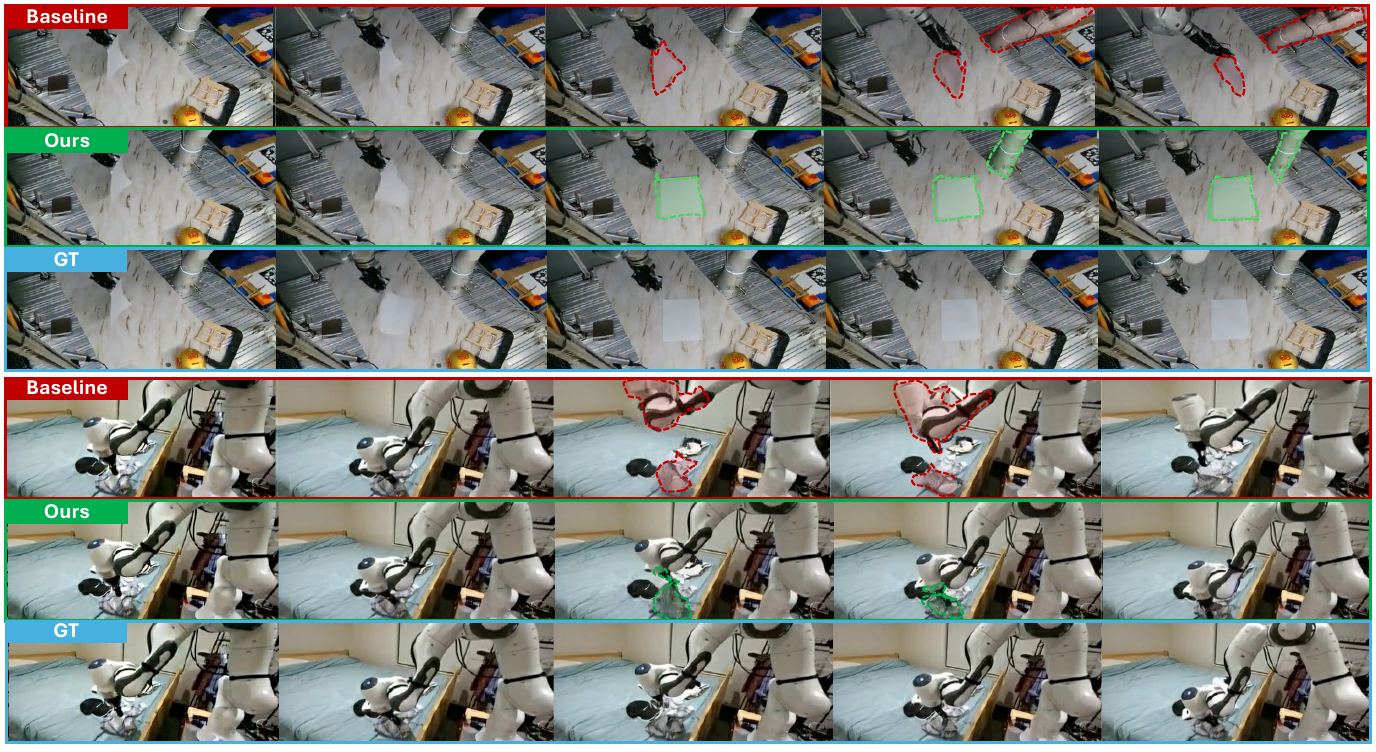}
    \caption{Qualitative comparison in two representative zero-shot, long-horizon scenarios from the test set: putting down a tissue and picking up a cloth. Each sequence represents a single-chunk prediction lasting approximately 4 seconds (64 frames). Compared to the action-guided I2V-baseline 14B model, our model demonstrates significantly better physical simulation of object deformations and gripper traces under long-horizon control, highlighting its superior generalization capabilities in unseen environments.}
  \label{fig:vis_wm}
\end{figure*}

\noindent\textbf{Evaluation of Control Protocols.} We further evaluate RoboInter-W under two distinct settings defined in our methodology: the \textit{oracle-control protocol} (using ground-truth annotations) and the \textit{planner-control protocol} (using autonomously generated representations). As detailed in Table~\ref{tab:protocols}, while the inevitable noise from the planner introduces a slight performance drop compared to the oracle upper bound, the generated representations successfully preserve essential geometric and temporal constraints. Consequently, the \textit{planner-control} formulation consistently and decisively outperforms the purely action-conditioned baselines across both 1.3B and 14B scales (e.g., yielding 20.17 vs. 18.26 PSNR on the 14B model). Ultimately, our framework \textbf{\textit{exhibits high robustness to imperfect priors}}; even autonomously generated intermediate representations provide sufficient structural scaffolding to surpass action-only simulations in practical deployment.

\noindent\textbf{Qualitative Analysis on RoboInter-World.} As illustrated in Figure.~\ref{fig:vis_wm}, we compare the performance of the baseline model with ours in long-horizon scenarios involving complex motions. It is evident that relying solely on abstract action sequences or language instructions is insufficient for the baseline model to achieve controllable motion simulation; the robotic arm either remains stagnant or exhibits unnatural structural distortions. Furthermore, the physical simulation of deformable objects is particularly poor, stemming from the limited scale of the training data and the model's limited generalization. In contrast, our approach explicitly introduces intermediate representations by leveraging the strong generalization of \textit{RoboInter-VLM}. The trace paths effectively constrain future motion control, concurrently, sampled object points enable more accurate simulation of rigid-body kinematics and flexible-object deformations, thereby significantly enhancing the physical realism.

\input{tables/wm_vla}

\subsection{World Model guides VLA Executor}
As shown in Table~\ref{tab:vla_accuracy}, relying solely on current observations, i.e., \textbf{VLA Baseline (No WM)}, leads to limited manipulation performance, as the model lacks explicit prediction and planning over future physical execution details. Conditioning the policy on future latent features predicted by a standard world model baseline (\textbf{VLA + I2V-Baseline}) provides only marginal improvement and even degrades performance under stricter thresholds (e.g., decreasing from 22.09\% to 21.07\% at Acc@0.03 for Step 55K), suggesting that unstructured predictive latent features may introduce harmful noise into action generation. In contrast, when conditioned on latent features predicted by our RoboInter-World (\textbf{VLA + RoboInter-W}), the policy achieves consistent and substantial gains across all evaluation thresholds, significantly narrowing the gap toward the \textbf{Oracle (GT)} upper bound, which provides the ground-truth latent features. These results demonstrate that \textbf{\textit{the latent representations produced by RoboInter-World are not merely visually plausible; they encode actionable physical priors that effectively enhance downstream manipulation performance.}}

%% file: tables/main_general_benchmark.tex
\begin{table}[t]
  \centering
  \scriptsize
  \caption{\textbf{Performance comparison on third-party benchmarks.} Including \textbf{Embodied}, \textbf{Grounding}, and \textbf{General} benchmarks for general VLMs (upper) and embodied VLMs (lower).}
  \renewcommand{\arraystretch}{1.1}
  \setlength{\tabcolsep}{1.9pt}
  \begin{tabularx}{0.96\linewidth}{l|*{3}{c}|*{3}{c}|*{6}{c}}
    \toprule
    \multirow{2}{*}{\textbf{Model Name}}
      & \multicolumn{3}{c|}{\textbf{Embodied}}
      & \multicolumn{3}{c|}{\textbf{Grounding}}
      & \multicolumn{6}{c}{\textbf{General}} \\ 
    \cmidrule(lr){2-4}\cmidrule(lr){5-7}\cmidrule(lr){8-13}
      & \makecell{\textbf{Where-}\\\textbf{2place$\uparrow$}}
      & \makecell{\textbf{RoboRefIt-}\\\textbf{test$\uparrow$}}
      & \makecell{\textbf{Robo-}\\\textbf{VQA$\uparrow$}}
      & \makecell{\textbf{Refcoco-}\\\textbf{g-val$\uparrow$}}
      & \makecell{\textbf{Refcoco+}\\\textbf{val$\uparrow$}}
      & \makecell{\textbf{Refcoco-}\\\textbf{val$\uparrow$}}
      & \makecell{\textbf{Text-}\\\textbf{VQA$\uparrow$}}
      & \makecell{\textbf{CO-}\\\textbf{CO$\uparrow$}}
      & \makecell{\textbf{OCR-}\\\textbf{bench$\uparrow$}}
      & \textbf{MME$\uparrow$}
      & \makecell{\textbf{MM-}\\\textbf{VET$\uparrow$}}
      & \textbf{POPE$\uparrow$} \\
    \midrule
    InternVL3-1B              &  2.65\% &  7.0\% & 30.5  & 79.8\% & 73.2\% & 83.0\% & 75.1 & 23.7 & 798 & 1907 & 58.9\% & 90.7\% \\
    InternVL3-2B              &  1.86\% & 27.5\% & 27.7  & 87.6\% & 84.0\% & 85.8\% & 77.0 & \textbf{27.9} & 835 & 2186 & 62.2\% & 89.6\% \\
    InternVL3-8B              &  1.95\% & 27.7\% & 27.9  & \textbf{89.6\%} & \textbf{88.2\%} & \textbf{92.5\%} & 80.2 & 26.5 & \underline{880} & \textbf{2410} & \textbf{81.3\%} & \underline{91.1\%} \\
    QwenVL2.5-3B              & 11.8\% & 68.9\% & 37.6  & 85.2\% & 82.4\% & 89.1\% & 79.3 & 15.7 & 797 & 2175 & 61.8\% & 85.9\% \\
    QwenVL2.5-7B              & 18.9\% & 75.8\% & 38.4  & 87.2\% & 84.2\% & 90.2\% & \textbf{84.9} & 15.0 & 864 & 2306 & 67.1\% & 85.9\% \\
    LLaVA-OV-7B               &  7.9\%  & 10.4\% & 31.4  & 71.9\% & 69.7\% & 73.8\% & 71.1 &  8.4 & \textbf{882} & \underline{2307} & \underline{67.3\%} & 86.4\% \\
    \midrule
    RoboBrain-2.0-3B        & 59.8\% & 30.9\% & 30.6  & 55.0\% & 51.5\% & 50.9\% & 81.0 & \underline{27.2} & 811 & 2126 & 59.4\% & 88.1\% \\
    RoboBrain-2.0-7B          & 63.6\% &  8.8\% & 31.6  & 62.9\% & 70.1\% & 76.1\% & 75.9 & 25.2 & 857 & 2076 & 61.4\% & 86.2\% \\
    \rowcolor{gray!17}
    RoboInter-Qwen-3B         & 58.3\% & 80.0\% & 43.3  & 87.9\% & 85.8\% & 89.5\% & 78.9 & 15.6 & 787 & 2180 & 61.0\% & 90.5\% \\
    \rowcolor{gray!17}
    RoboInter-Qwen-7B         & \underline{65.8\%} & \underline{85.6\%} & \underline{74.4}  & \underline{88.4\%} & \underline{86.6\%} & \underline{91.5\%} & \underline{83.0} & 15.9 & 832 & 2281 & 62.3\% & \textbf{91.4\%} \\
    \rowcolor{gray!17}
    RoboInter-LLaVAOV-7B     & \textbf{66.3\%} & \textbf{89.3\%} & \textbf{74.5}  & 87.3\% & 84.2\% & 91.3\% & 72.2 & 15.8 & 725 & 2217 & 61.4\% & 90.4\% \\
    \bottomrule
  \end{tabularx}
  \label{tab:multi_task_performance_reordered}
\end{table}

%% file: tables/main_RoboInterBenchmark.tex
\begin{table}[t]
  \centering
  \scriptsize
    \caption{\textbf{Results of RoboInter-VQA \textit{spatial} and \textit{temporal} benchmark.} \textbf{G.D.} means grounding, \textbf{A.F.} denotes Affordance. Spatial generation uses ACC@IOU$>$0.1 (\%↑), multiple choice and T/F use ACC (\%↑). For temporal, Trace uses DTW (↓), other metrics use ACC or average BLEU (↑).}
    \label{tab:robointer_spatial_temporal_revised}
  \renewcommand{\arraystretch}{1.0}
  \setlength{\tabcolsep}{1.6pt}
  \begin{tabularx}{0.99\linewidth}{l|ccccccc|ccccc}
    \toprule
    \multirow{3}{*}{\textbf{Model Name}}
      & \multicolumn{7}{c|}{\textbf{RoboInter-VQA Spatial}}
      & \multicolumn{5}{c}{\textbf{RoboInter-VQA Temporal}} \\
    \cmidrule(lr){2-8}\cmidrule(lr){9-13}
      & \multicolumn{4}{c}{\textbf{Generation}}
      & \multicolumn{2}{c}{\textbf{Multiple Choice}}
      & \multicolumn{1}{c|}{\textbf{T/F}}
      & \multicolumn{2}{c}{\textbf{Generation}}
      & \multicolumn{2}{c}{\textbf{Multiple Choice}}
      & \multicolumn{1}{c}{\textbf{T/F}} \\
    \cmidrule(lr){2-5}\cmidrule(lr){6-7}\cmidrule(lr){8-8}
    \cmidrule(lr){9-10}\cmidrule(lr){11-12}\cmidrule(lr){13-13}
      & \makecell{\textbf{Object}\\\textbf{G.D.$\uparrow$}}
      & \makecell{\textbf{Grasp}\\\textbf{A.F.$\uparrow$}}
      & \makecell{\textbf{Place}\\\textbf{A.F.$\uparrow$}}
      & \makecell{\textbf{Gripper}\\\textbf{G.D.$\uparrow$}}
      & \makecell{\textbf{Grasp}\\\textbf{Pose$\uparrow$}}
      & \makecell{\textbf{Grouding}\\\textbf{Choice$\uparrow$}}
      & \textbf{Contact$\uparrow$}
      & \makecell{\textbf{Trace}$\downarrow$}
      & \makecell{\textbf{Task}\\\textbf{Planning}$\uparrow$}
      & \makecell{\textbf{Visual}\\\textbf{Trace}$\uparrow$}
      & \makecell{\textbf{Planning}\\\textbf{Choice}$\uparrow$}
      & \makecell{\textbf{Task}\\\textbf{Planning}$\uparrow$} \\
    \midrule

    QwenVL2.5-3B              & 46.6\% & 12.2\% & 34.1\% &  6.1\% & 21.1\% & 21.9\% & 50.9\% & 2712 & 20.3 & 37.5\% & 60.0\% & 59.7\% \\
    QwenVL2.5-7B              & 51.2\% & 14.7\% & 38.2\% & 10.2\% & 27.3\% & 25.7\% & 52.5\% & 1702 & 22.4 & 39.0\% & 64.5\% & 60.5\% \\
    InternVL3-1B              &  7.8\% &  2.3\% &  8.3\% &  1.2\% & 24.8\% & 25.9\% & 50.4\% &  --   & 10.5 & 28.9\% & 54.9\% & 55.8\% \\
    InternVL3-2B              & 20.6\% &  3.1\% & 17.9\% &  1.9\% & 25.5\% & 27.3\% & 50.1\% &  --   &  7.7 & 35.2\% & 59.3\% & 59.4\% \\
    InternVL3-8B              & 32.7\% &  5.9\% & 28.2\% &  3.5\% & 25.1\% & 31.1\% & 52.9\% & 1035 &  8.1 & 34.0\% & 71.5\% & 60.0\% \\
    Llava-OV-7B               & 25.8\% &  5.5\% & 23.7\% &  1.6\% & 24.5\% & 31.9\% & 54.4\% &  --   & 11.0 & 37.7\% & 44.9\% & 63.5\% \\
    \midrule
            GPT4o-mini                &  6.8\% &  --     &  7.2\% &  1.1\% & 10.9\% & 16.8\% & 53.6\% & 1736 & 14.7 & 28.4\% & 66.6\% & 63.9\% \\
    Gemini-2.5-flash          &  1.7\% &  --     &  1.2\% &   --    & 32.7\% & 69.4\% & 65.5\% &  --   &  --  & 49.4\% &  --   &  --   \\
    \midrule
    RoboBrain-2.0-3B          & 15.2\% &  --     &  --     &  2.8\% & 25.5\% & 26.5\% & 50.4\% &  595 & 16.0 & 29.7\% & 48.2\% & 46.8\% \\
    RoboBrain-2.0-7B          &  --     &  --     &  --     &  2.5\% & 23.3\% & 21.5\% & 49.2\% &  541 & 15.3 & 29.5\% & 57.8\% & 46.4\% \\
    \rowcolor{gray!14}
    RoboInter-Qwen-3B         & 76.1\% & 34.9\% & 52.7\% & 61.6\% & 74.0\% & 73.4\% & 75.2\% &  332 & 61.2 & 78.8\% & 82.2\% & 88.7\% \\
    \rowcolor{gray!14}
    RoboInter-Qwen-7B         & 75.1\% & 37.8\% & \textbf{56.9\%} & 62.0\% & \textbf{76.1\%} & 75.7\% & 75.6\% &  323 & \textbf{63.4} & \textbf{81.9\%} & \textbf{86.5\%} & \textbf{93.0\%} \\
    \rowcolor{gray!14}
    RoboInter-LlavaOV-7B      & \textbf{82.9\%} & \textbf{46.3\%} & 55.1\% & \textbf{70.1\%} & 74.1\% & \textbf{79.7\%} & \textbf{76.3\%} &  \textbf{299} & 62.7 & \textbf{81.9\%} & 81.8\% & 83.9\% \\
    \bottomrule
  \end{tabularx}
\end{table}

%% file: tables/tab_OpenL_OH.tex
\begin{figure}[t]
  \centering

  \begin{minipage}[t]{0.5\linewidth}
    \vspace{1pt}
    \centering
    \captionof{table}{\textbf{Open-loop evaluation in In-the-Wild setting}. We report OLS with different error thresholds (@0.1 to @0.01) and the mean value.}
    \renewcommand{\arraystretch}{1.2}
    \setlength{\tabcolsep}{3pt}
    \scriptsize
    \begin{tabularx}{\linewidth}{c|cccc|c}
    \toprule
    \textbf{Method} & \multicolumn{4}{c|}{\textbf{Open-Loop Score (OLS)}} & \textbf{mOLS} \\
    \cmidrule(lr){2-5}
                    & @0.1 & @0.05 & @0.03 & @0.01 & - \\
    \midrule
    \modify{VLA-OS}    & \modify{0.6180} & \modify{0.3905} & \modify{0.1928} & \modify{0.0129} & \modify{0.3035}\\
    Vanilla                  & 0.6793 & 0.3608 & 0.1753 & 0.0189 & 0.3086\\\rowcolor{gray!15}
    RoboInter-IC-E2E           & 0.6984 & 0.3810 & 0.1873 & 0.0204 
    & 0.3218 \\\rowcolor{gray!15}
    RoboInter-EC-E2E           & 0.7049 & 0.3930 & 0.2066 & 0.0314 &
    0.3340 \\
    \midrule
    QwenVL+Executor          & 0.6749 &  0.3582 & 0.1777 & 0.0298 & 0.3102  \\\rowcolor{gray!15}
    RoboInter-Te-Modular        & 0.7124 & 0.4133 & 0.2332 & 0.0584 &
    0.3543 \\\rowcolor{gray!15}
    RoboInter-Im-Modular      & 0.7056 & 0.4029 & 0.2240 & 0.0430 &
    0.3439 \\
    \modify{Oracle+VLA-OS}  & \modify{0.7260} & \modify{0.4928} & \modify{0.2734} & \modify{0.0200} & \modify{0.3780} \\
    Oracle+Executor  & 0.7511 & 0.4640 & 0.2705 & 0.0587 &
    0.3861 \\
    \bottomrule
    \end{tabularx}
    \label{tab:OpenL_OH}
  \end{minipage}
  \hfill
  \begin{minipage}[t]{0.49\linewidth}
    \vspace{0pt}
    \centering
    \includegraphics[width=\linewidth]{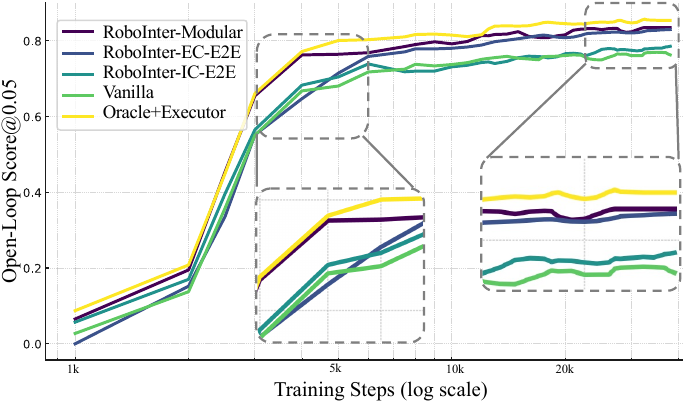}
    \caption{\textbf{Open-loop evaluation in TableTop setting}. We show the curve of OLS@0.05 from 1k to 40k training steps. We mainly report the five variances of RoboInter-VLA.}
    \label{fig:OpenL_TT}
  \end{minipage}

  \vspace{-10pt}
\end{figure}

%% file: tables/wm_main.tex
\begin{figure}[t]
  \begin{minipage}[t]{0.49\linewidth}
    \centering
    \captionof{table}{\textbf{Main quantitative benchmarking of world model predictions.} Oracle visual generation quality across various model architectures and fine-tuning strategies is shown.}
    \label{tab:main_results}
    \renewcommand{\arraystretch}{1.1}
    \setlength{\tabcolsep}{2pt}
    \scriptsize
    \resizebox{0.98\columnwidth}{!}{
    \begin{tabular}{@{}l|c|ccccc@{}}
    \toprule
    \textbf{Model} & \textbf{Control} & \textbf{PSNR$\uparrow$} & \textbf{SSIM$\uparrow$} & \textbf{LPIPS$\downarrow$} & \textbf{MAE$\downarrow$} & \textbf{MSE$\downarrow$} \\\midrule
    \multicolumn{7}{c}{\textbf{1.3B Parameter}} \\ \midrule
    T2V-Baseline(Full-FT) & Action & 18.21 & 0.739 & 0.176 & 0.072 & 0.022 \\
    RoboInter-W(Full-FT) & Action & 19.08 & 0.770 & 0.148 & 0.063 & 0.029 \\
    \rowcolor{gray!15}
    RoboInter-W(Full-FT) & Inter & 20.00 & 0.795 & 0.125 & 0.055 & 0.015 \\ \midrule
    T2V-Baseline(LoRA) & Action & 18.49 & 0.763 & 0.163 & 0.067 & 0.022 \\
    RoboInter-W(LoRA) & Action & 19.08 & 0.770 & 0.148 & 0.063 & 0.029 \\\rowcolor{gray!15}
    RoboInter-W(LoRA) & Inter & 20.43 & 0.800 & 0.114 & 0.052 & 0.014 \\ \midrule
    \multicolumn{7}{c}{\textbf{14B Parameter}} \\ \midrule
    I2V-Baseline(LoRA) & Action & 19.43 & 0.770 & 0.147 & 0.063 & 0.017 \\
    T2V-Baseline(LoRA) & Action & 19.36 & 0.780 & 0.141 & 0.059 & 0.018 \\
    RoboInter-W(LoRA) & Action & 18.26 & 0.750 & 0.171 & 0.072 & 0.023 \\
    \rowcolor{gray!15}
    RoboInter-W(LoRA) & Inter & 21.05 & 0.810 & 0.102 & 0.047 & 0.013 \\ \bottomrule
    \end{tabular}}
  \end{minipage}
  \hfill
  \begin{minipage}[t]{0.5\linewidth}
    \centering
    \captionof{table}{\textbf{Impact of context and prediction horizons (14B LoRA).} We compare the predictive fidelity of raw action conditioning versus our intermediate representations (\textit{Seg+Trace}) across varying history ($H$) and prediction ($P$) lengths. Note that $H$ and $P$ denote the number of latent frames after 3D VAE encoding, which corresponds to $4\times$ the raw video frames due to temporal compression.}
    \label{tab:horizons}
    \renewcommand{\arraystretch}{1.1}
    \setlength{\tabcolsep}{2pt}
    \scriptsize
    \resizebox{0.98\columnwidth}{!}{
    \begin{tabular}{@{}l|c|l|ccc|c@{}}
    \toprule
    \multirow{2}{*}{\textbf{Model}} & \multirow{2}{*}{\textbf{H/P}} & \multirow{2}{*}{\textbf{Control}} & \multicolumn{3}{c|}{\textbf{Metrics}} & \multirow{2}{*}{\textbf{$\Delta$ PSNR}} \\ \cmidrule(lr){4-6}
     & & & \textbf{PSNR$\uparrow$} & \textbf{SSIM$\uparrow$} & \textbf{LPIPS$\downarrow$} & \\ \midrule
     
     RoboInter-W Baseline & 4/3 & Action & 24.87 & 0.878 & 0.054 & \\
     \textbf{RoboInter-W} & 4/3 & \textbf{Seg+Trace} & \textbf{25.00} & \textbf{0.881} & \textbf{0.049} & \multirow{-2}{*}{+0.13} \\ \midrule
     
     RoboInter-W Baseline & 1/3 & Action & 26.20 & 0.886 & 0.051 & \\
     \textbf{RoboInter-W} & 1/3 & \textbf{Seg+Trace} & \textbf{27.70} & \textbf{0.899} & \textbf{0.041} & \multirow{-2}{*}{+1.50} \\ \midrule
     
     RoboInter-W Baseline & 4/16 & Action & 18.26 & 0.750 & 0.171 & \\
     \textbf{RoboInter-W} & 4/16 & \textbf{Seg+Trace} & \textbf{21.05} & \textbf{0.810} & \textbf{0.102} & \multirow{-2}{*}{\textbf{+2.79}} \\ \bottomrule
     
    \end{tabular}
    }
  \end{minipage}
  \vspace{-10pt}
\end{figure}

%% file: tables/wm_abl_modality.tex
\begin{table}[t]

\end{table}

\begin{figure}[t]
  \begin{minipage}[t]{0.49\linewidth}
    \centering
    \captionof{table}{Ablation study on control modalities across model scales. We focus on action, segmentation masks, and 2D trace conditions.}
    \label{tab:ablation}
    \renewcommand{\arraystretch}{1.1}
    \setlength{\tabcolsep}{4pt}
    \scriptsize
    \resizebox{0.98\columnwidth}{!}{
    \begin{tabular}{@{}l|lccccc@{}}
    \toprule
    \multirow{2}{*}{\textbf{Model}} & \multirow{2}{*}{\textbf{Control}} & \multicolumn{5}{c}{\textbf{Metrics}} \\ \cmidrule(l){3-7}
     & & \textbf{PSNR$\uparrow$} & \textbf{SSIM$\uparrow$} & \textbf{LPIPS$\downarrow$} & \textbf{MAE$\downarrow$} & \textbf{MSE$\downarrow$} \\ \midrule
    \multicolumn{7}{c}{\textbf{1.3B Parameter Models (LoRA)}} \\ \midrule
    T2V-Baseline & Action & 18.49 & 0.763 & 0.163 & 0.067 & 0.022 \\
    RoboInter-W & Action & 19.08 & 0.770 & 0.148 & 0.063 & 0.029 \\
    \rowcolor{gray!15}
    RoboInter-W & Seg & 20.38 & 0.804 & 0.113 & 0.052 & 0.014 \\
    \rowcolor{gray!15}
    RoboInter-W & Trace & 20.01 & 0.791 & 0.125 & 0.056 & 0.016 \\
    \rowcolor{gray!15}
    RoboInter-W & Seg+Trace & 20.43 & 0.800 & 0.114 & 0.052 & 0.014 \\ \midrule
    
    \multicolumn{7}{c}{\textbf{14B Parameter Models (LoRA)}} \\ \midrule
    I2V-Baseline & Action & 19.43 & 0.770 & 0.147 & 0.063 & 0.017 \\
    RoboInter-W & Action & 18.26 & 0.750 & 0.171 & 0.072 & 0.023 \\
    \rowcolor{gray!15}
    RoboInter-W & Seg & 20.44 & 0.807 & 0.109 & 0.049 & 0.013 \\
    \rowcolor{gray!15}
    RoboInter-W & Trace & 20.42 & 0.800 & 0.119 & 0.053 & 0.015 \\
    \rowcolor{gray!15}
    RoboInter-W & Seg+Trace & 21.05 & 0.810 & 0.102 & 0.047 & 0.013 \\ \bottomrule
    \end{tabular}
    }
  \end{minipage}
  \hfill
    \begin{minipage}[t]{0.5\linewidth}
        \centering
        \captionof{table}{\textbf{Evaluation of different control protocols across model scales.} Predictive fidelity of RoboInter-W under the \textit{planner-control protocol} versus the \textit{oracle-control protocol} (upper bound) and action-only baselines.}
        \label{tab:protocols}
        \renewcommand{\arraystretch}{1.1}
        \setlength{\tabcolsep}{2pt}
        \scriptsize
        \resizebox{\columnwidth}{!}{
        \begin{tabular}{@{}l|c|c|ccccc@{}}
        \toprule
        \multirow{2}{*}{\textbf{Model}} & \multirow{2}{*}{\textbf{Control}} & \multirow{2}{*}{\textbf{Protocol}} & \multicolumn{5}{c}{\textbf{Metrics}} \\ \cmidrule(l){4-8}
         & & & \textbf{PSNR$\uparrow$} & \textbf{SSIM$\uparrow$} & \textbf{LPIPS$\downarrow$} & \textbf{MAE$\downarrow$} & \textbf{MSE$\downarrow$} \\ \midrule
         
         \multicolumn{8}{c}{\textbf{1.3B Parameter Models (LoRA)}} \\ \midrule
        T2V-Baseline & Action & -- & 18.49 & 0.763 & 0.163 & 0.067 & 0.022 \\
        RoboInter-W & Action & -- & 19.08 & 0.770 & 0.148 & 0.063 & 0.029 \\
        \rowcolor{gray!15}
        RoboInter-W & \textit{Inter} & Planner-Control & 19.95 & 0.780 & 0.128 & 0.061 & 0.021 \\
        \rowcolor{gray!15}
        RoboInter-W & \textit{Inter} & Oracle-Control & 20.43 & 0.800 & 0.114 & 0.052 & 0.014 \\ \midrule
        
         \multicolumn{8}{c}{\textbf{14B Parameter Models (LoRA)}} \\ \midrule
        I2V-Baseline & Action & -- & 19.43 & 0.770 & 0.147 & 0.063 & 0.017 \\
        RoboInter-W & Action & -- & 18.26 & 0.750 & 0.171 & 0.072 & 0.023 \\
        \rowcolor{gray!15}
        RoboInter-W & \textit{Inter} & Planner-Control & 20.17 & 0.780 & 0.134 & 0.058 & 0.017 \\
        \rowcolor{gray!15}
        RoboInter-W & \textit{Inter} & Oracle-Control & 21.06 & 0.810 & 0.102 & 0.047 & 0.013 \\ \bottomrule
        
        \end{tabular}
        }
  \end{minipage}
  \vspace{-10pt}
\end{figure}

%% file: tables/wm_vla.tex
\begin{table}[t]
\centering
\caption{\textbf{World-guided VLA action accuracy (\%).} We evaluate the impact of predicted latent representations on a downstream VLA model across different training steps. \textit{OLS@X} denotes the open-Loop score of predicted actions within a normalized distance threshold.}
\label{tab:vla_accuracy}
\renewcommand{\arraystretch}{1.1}
\setlength{\tabcolsep}{3pt}
\scriptsize
\resizebox{0.68\columnwidth}{!}{
\begin{tabular}{@{}l|l|cccc@{}}
\toprule
\multirow{2}{*}{\textbf{Step}} & \multirow{2}{*}{\textbf{VLA Conditioning Source}} & \multicolumn{4}{c}{\textbf{Open-Loop Score (\%) $\uparrow$}} \\ \cmidrule(l){3-6}
 & & \textbf{OLS@0.03} & \textbf{OLS@0.05} & \textbf{OLS@0.10} & \textbf{OLS@0.20} \\ \midrule
 
 \multirow{4}{*}{35K} 
 & VLA Baseline (No WM) & 20.37 & 34.30 & 53.04 & 63.44 \\
 & VLA + I2V-Baseline (Pred) & 19.71 & 33.72 & 53.91 & 64.27 \\
 & \textbf{VLA + RoboInter-W (Pred)} & \textbf{21.42} & \textbf{35.23} & \textbf{54.54} & \textbf{64.42} \\
 & VLA + Oracle (GT) & 21.78 & 35.75 & 54.86 & 64.96 \\ \midrule
 
 \multirow{4}{*}{55K} 
 & VLA Baseline (No WM) & 22.09 & 35.74 & 54.26 & 64.33 \\
 & VLA + I2V-Baseline (Pred) & 21.07 & 34.85 & 55.01 & 65.68 \\
 & \textbf{VLA + RoboInter-W (Pred)} & \textbf{22.17} & \textbf{35.97} & \textbf{55.39} & \textbf{65.55} \\
 & VLA + Oracle (GT) & 22.80 & 36.42 & 55.60 & 65.41 \\ \bottomrule

\end{tabular}
}
\end{table}

%% file: section/6_conclusion.tex
\section{Conclusion}
In this work, we presented the \textit{RoboInter1.5} Manipulation Suite, a unified platform designed to advance research on intermediate representations for the plan-then-execute paradigm. As its core, \textit{RoboInter-Data} provides over 230k episodes with dense, per-frame annotations, establishing a new standard of scale and quality for real-world manipulation datasets. Built upon this foundation, \textit{RoboInter-VQA} systematically benchmarks the embodied understanding and generation capabilities of \textit{RoboInter-VLMs} across complex spatiotemporal tasks. Bridging the gap from static understanding to physical interaction, \textit{RoboInter-VLA} integrates these geometric priors into both modular and end-to-end control frameworks, enabling a principled investigation of how intermediate representations drive superior execution performance. Finally, extending these capabilities into generative simulation, \textit{RoboInter-World} serves as a robust physical simulator, leveraging these actionable representations to accurately forecast future environmental dynamics.

\clearpage
\newpage

%% file: section/supp.tex
\section{Appendix}
\label{sec_supp}


\subsection{Contributors}
\label{sec:contributors}

Ziqin Wang$^{1, 3, *}$, Hao Li$^{2, 3, *}$, Weijun Wang$^{1}$, Junhao Cai$^{3}$, Jia Zeng$^{3}$, Yilun Chen$^{3}$, Jiangmiao Pang$^{3}$, Si Liu$^{1}$

{\let\thefootnote\relax
\footnotetext{$^{*}$\,Equal core contributions, ordered by coin toss. $^1$Beihang University, $^2$University of Science and Technology of China, $^3$Shanghai Artificial Intelligence Laboratory.}}

\subsection{Statements}

\textit{RoboInter1.5} is an extended version of our previously published technical report, \textit{RoboInter: A Holistic Intermediate Representation Suite Towards Robotic Manipulation}~\citep{li2026robointer}. The present work builds directly upon the framework, methodology, and experimental foundation established in the earlier version of \textit{RoboInter}. Accordingly, substantial portions of the main text and supplementary material are inherited from or adapted from the original report, while the newly introduced components, experiments, and analyses constitute the primary extensions presented in \textit{RoboInter1.5}. For detailed appendix content, please refer to RoboInter1.0~\citep{li2026robointer}.

\subsection{Additional Real-world ID and OOD Validation}
\begin{figure}[h]
    \centering
    \includegraphics[width=\linewidth]{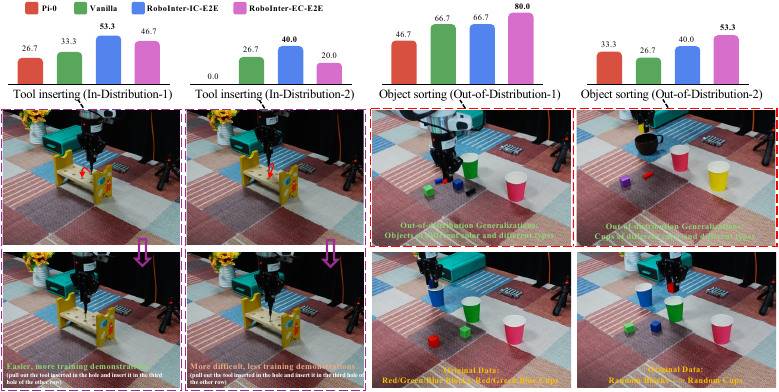}
    \caption{\modify{\textbf{Additional real-world ID and OOD validation.} (Left) \textit{Tool Inserting}: a precision ID task requiring accurate contact handling and slot alignment. (Right) \textit{Object Sorting}: an OOD task that tests language-guided generalization using novel objects and containers.}}
    \label{fig:insert_sort_real_world}
\end{figure}

To further disentangle in-distribution (ID) and out-of-distribution (OOD) generalization, we introduce two additional real-world tasks that emphasize different aspects of control and reasoning:

\begin{itemize}[leftmargin=*]

\item \modify{\textbf{Tool Inserting (ID-oriented).}}  
\modify{A precision-controlled manipulation task that evaluates the model’s ability to fit fine-grained in-distribution actions. The robot must pull a metal tool out of a $1.5\,\text{cm} \times 1.5\,\text{cm}$ slot and re-insert it into another slot of the same size. Two ID variants with different initial positions and different training demonstrations are used, each requiring accurate contact handling.}

\item \modify{\textbf{Object Sorting (OOD-oriented).}}  
\modify{A generalization task in which the robot must place objects into their corresponding target containers. Training demonstrations include only red, green, and blue objects and cups with simple pick-and-place motions. The OOD setting introduces novel objects and cups with novel colors, shapes, or types, assessing whether models can generalize sorting behaviors according to language instructions beyond the training distribution.}
\end{itemize}

The results in Figure~\ref{fig:insert_sort_real_world} show complementary strengths of the two variants: EC-E2E achieves stronger OOD performance owing to its explicit reasoning, whereas IC-E2E exhibits superior ID robustness. Overall, both variants outperform the Vanilla and $\pi_0$ baselines, demonstrating that intermediate representations significantly benefit both ID precision and OOD generalization.

\subsection{Additional Qualitative results on RoboInter-VLA}
\begin{figure*}
  \centering
  \includegraphics[width=0.75\linewidth]{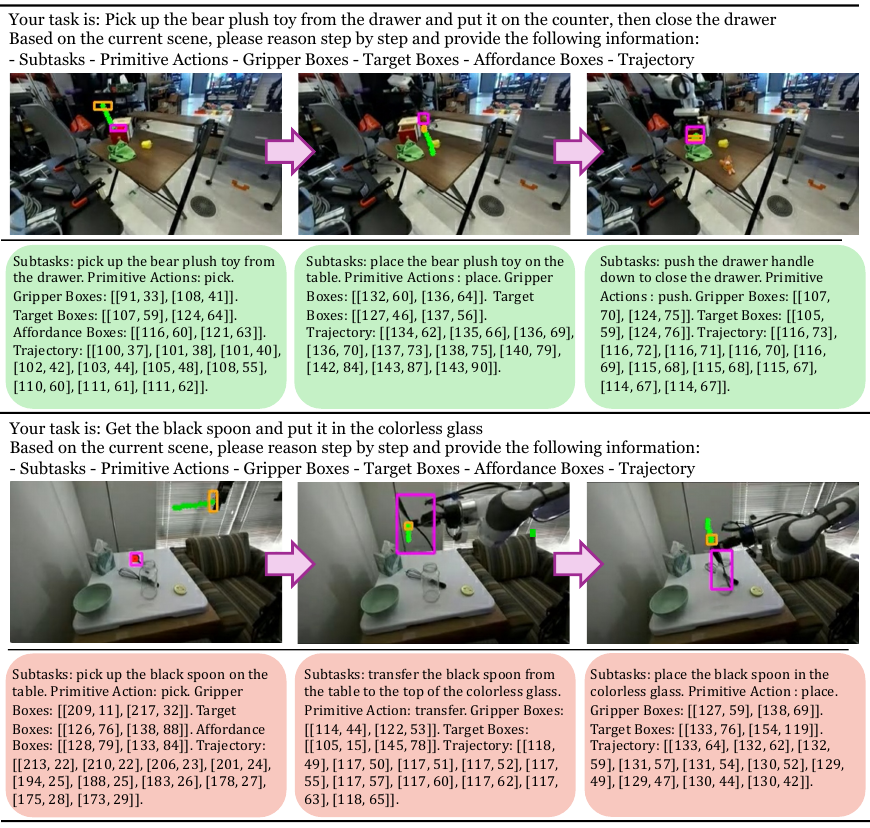}
  \caption{\textbf{Chain of thought of RoboInter-VLA}. Two representative examples (used in RoboInter-EC-E2E and RoboInter-Modular) are shown. The upper example involves a sequence of manipulation behaviors, including picking, placing, and pushing, while the lower one focuses on a pick-and-place task. RoboInter-VLA predicts and utilizes structured chain-of-thought representations consisting of subtasks, primitive skills, gripper boxes, target object boxes, affordance boxes, and traces.}
  \label{fig:vla}
\end{figure*}

Figure.\ref{fig:vla} shows that, in in-the-wild open-loop scenarios, RoboInter-VLA first predicts structured chain-of-thought representations before generating actions. Such intermediate reasoning provides the VLA model with a richer perception of the environment and task semantics, thereby improving its generalization ability and execution accuracy.

